\pdfoutput=1
\documentclass[11pt,a4paper]{article}
\usepackage[hyperref]{emnlp2020}
\usepackage{times}
\usepackage{latexsym}

\newcommand{\sectioncolor}{black}

\usepackage{microtype}

\usepackage{tikz}
\usetikzlibrary{snakes}
\usepackage{tabularx}
\usepackage{booktabs}
\usepackage{makecell}
\usepackage{multirow}
\usepackage{amssymb}
\usepackage{amsmath}
\usepackage{subcaption}
\usepackage{xspace}
\usepackage{mdframed}
\usepackage[utf8]{inputenc}

\usepackage{paralist}

\renewenvironment{enumerate}[1]{\begin{compactenum}#1}{\end{compactenum}}

\usepackage{algorithm}
\usepackage[noend]{algpseudocode}

\newcolumntype{Y}{>{\centering\arraybackslash}X}

\aclfinalcopy % Uncomment this line for the final submission
\def\aclpaperid{1113} %  Enter the acl Paper ID here

\newcommand{\eg}{e.g.,~}
\newcommand{\ie}{i.e.,~}

\newcommand{\figref}[1]{Figure \ref{#1}}

\newcommand{\tabref}[1]{Table \ref{#1}}

\renewcommand{\O}{{\cal O}}

\newcommand{\pz}{\hphantom{0}}
\newcommand{\pzz}{\hphantom{00}}

\newcommand{\graphp}{\ensuremath{P}\xspace}
\newcommand{\graphr}{\ensuremath{R}\xspace}

\title{Room-Across-Room: Multilingual Vision-and-Language\\Navigation with Dense Spatiotemporal Grounding}

\author{Alexander Ku\textsuperscript{1}\thanks{\enspace First two authors contributed equally.} \quad Peter Anderson\textsuperscript{1}\footnotemark[1] \quad Roma Patel\textsuperscript{2} \quad Eugene Ie\textsuperscript{1} \quad Jason Baldridge\textsuperscript{1}\\
$^1$Google Research \quad  $^2$Brown University\\
\texttt{\normalsize{\{alexku, pjand, eugeneie, jridge\}@google.com romapatel@brown.edu }}}

\date{}

\begin{document}
\maketitle

\begin{abstract}
We introduce Room-Across-Room (RxR), a new Vision-and-Language Navigation (VLN) dataset. RxR is multilingual (English, Hindi, and Telugu) and larger (more paths and instructions) than other VLN datasets. It emphasizes the role of language in VLN by addressing known biases in paths and eliciting more references to visible entities. Furthermore, each word in an instruction is time-aligned to the virtual poses of instruction creators and validators. We establish baseline scores for monolingual and multilingual settings and multitask learning when including Room-to-Room annotations \cite{mattersim}. We also provide results for a model that learns from synchronized pose traces by focusing only on portions of the panorama attended to in human demonstrations. The size, scope and detail of RxR dramatically expands the frontier for research on embodied language agents in simulated, photo-realistic environments.
\end{abstract}

\section{Introduction}

Vision-and-Language Navigation (VLN) tasks require computational agents to mediate the relationship between language, visual scenes and movement. Datasets have been collected for both indoor \cite{mattersim,thomason:corl19,qi2019reverie} and outdoor \cite{chen2019touchdown,mehta2020retouchdown} environments; success in these is based on clearly-defined, objective task completion rather than language or vision specific annotations. These VLN tasks fall in the Goldilocks zone: they can be tackled -- but not solved -- with current methods, and progress on them makes headway on real world grounded language understanding.

\begin{figure}
\begin{center}
\includegraphics[width=.99\linewidth]{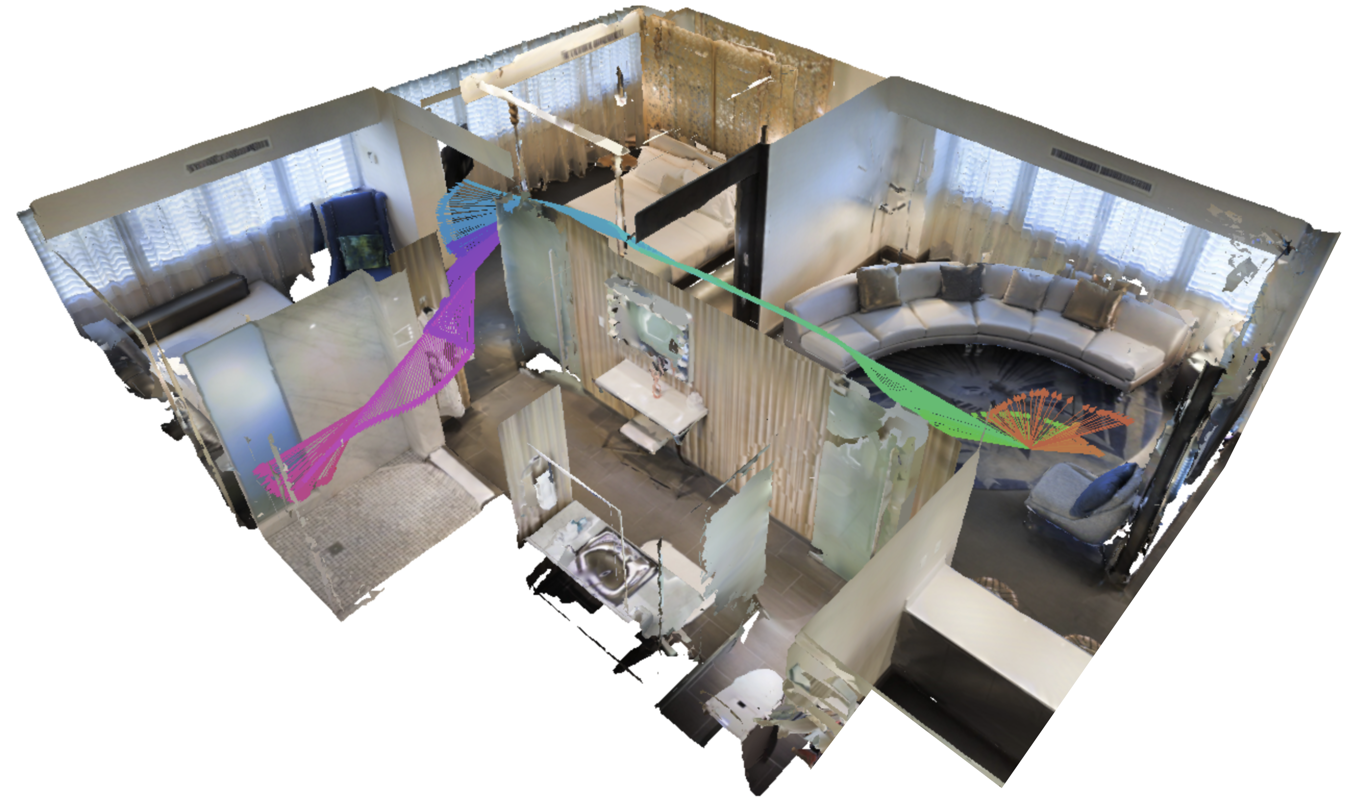}\\
\vspace{3mm}
\includegraphics[width=.95\linewidth]{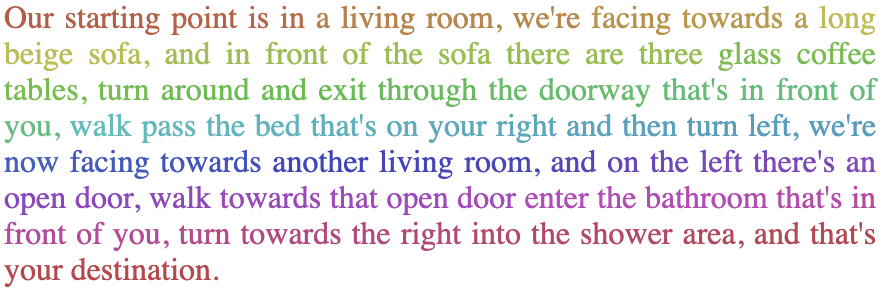}
\end{center}
\caption{RxR's instructions are densely grounded to the visual scene by aligning the annotator's virtual pose to their spoken instructions for navigating a path.}
\label{fig:teaser}
\end{figure}

We introduce Room-across-Room (RxR), a VLN dataset that addresses gaps in existing ones by (1) including more paths that (2) counter known biases in existing datasets, and (3) collecting an order of magnitude more instructions for (4) three languages (English, Hindi and Telugu) while (5) capturing annotators' 3D pose sequences. As such, RxR includes dense spatiotemporal grounding for every instruction, as illustrated in \figref{fig:teaser}.

We provide monolingual and multilingual baseline experiments using a variant of the Reinforced Cross-Modal Matching agent \cite{wang2018reinforced}. 
Performance generally improves by using monolingual learning, and by using RxR's follower paths as well as its guide paths. 
We also concatenate R2R and RxR annotations as a simple multitask strategy \cite{wang2020environmentagnostic}: the agent trained on both datasets obtains across the board improvements. 

RxR contains 126K instructions covering 16.5K sampled guide paths and 126K human follower demonstration paths. The dataset is available.\footnote{\href{https://github.com/google-research-datasets/RxR}{https://github.com/google-research-datasets/RxR}} We plan to release a test evaluation server, our annotation tool, and code for all experiments.
\section{Motivation}

A number of VLN datasets situated in photo-realistic 3D reconstructions of real locations contain human instructions or dialogue: R2R \cite{mattersim}, Touchdown \cite{chen2019touchdown,mehta2020retouchdown}, CVDN \cite{thomason:corl19} and REVERIE \cite{qi2019reverie}. RxR addresses shortcomings of these datasets---in particular, multilinguality, scale, fine-grained word grounding, and human follower demonstrations (\tabref{tab:datasets}). It also addresses path biases in R2R. More broadly, our work is also related to instruction-guided household task benchmarks such as ALFRED \cite{ALFRED20} and CHAI \cite{MisraBBNSA18}. These synthetic environments provide interactivity but are generally less diverse, less visually realistic and less faithful to real world structures than the 3D reconstructions used in VLN. 

\textbf{Multilinguality.}
The dominance of high resource languages is a pervasive problem as it is unclear that research findings generalize to other languages \cite{bender-2009-linguistically}. The issue is particularly severe for VLN. \citet{chen2011learning} translated($\sim$1K) English navigation instructions into Chinese for a game-like simulated 3D environment. Otherwise, all publicly available VLN datasets we are aware of have English instructions.

To enable multilingual progress on VLN, RxR includes instructions for three typologically diverse languages: English (en), Hindi (hi), and Telugu (te). The English portion includes instructions by speakers in the USA (en-US) and India (en-IN). Unlike \citet{chen2011learning} and like the TyDi-QA multilingual question answering dataset \cite{tydiqa}, RxR's instructions are not translations: all instructions are created from scratch by native speakers. This especially matters for VLN, as different languages encode spatial and temporal information in idiosyncratic ways--\eg how contact/support relationships are expressed \cite{munnich2001spatial}, frame of reference \cite{haun2011plasticity}, and how temporal accounts are expressed \cite{bender2014mapping}. 

\textbf{Scale.}
Embodied language tasks suffer from a relative paucity of training data; for VLN, this has led to a focus on data augmentation \cite{fried2018speaker,backtranslate2019}, pre-training \cite{wang2018reinforced,huang2019multi,li2019robust}, multi-task learning \cite{wang2020environmentagnostic} and better generalization through piece-wise curriculum design \cite{zhu2020bwalk}. To address this shortage, for each language RxR contains 14K paths with 3 instructions per path, for a total of 126K instructions and 10M words (based on whitespace tokenization). As illustrated in \tabref{tab:datasets}, this is \textit{an order of magnitude} larger than previous datasets. 

\begin{table}
\centering
\footnotesize
\setlength\tabcolsep{1.3pt}
\begin{tabularx}{\columnwidth}{Xcrrrcccc}
          & \multicolumn{4}{c}{Number of:}    &  & \multicolumn{3}{c}{Includes:}       \\
\cmidrule{2-5} \cmidrule{7-9}
          & Lang & Instruct & Words & Paths & & Text & Ground & Demos \\
CVDN      & 1         &       2K$^\dagger$       &  167K     &  7K & &   \checkmark     &            &       \\
R2R       & 1         &       22K\pz    &   625K    & 7K & &   \checkmark      &            &       \\
Touchdown & 1         &       9K\pz     &   1.0M    & 9K & &   \checkmark      &     \checkmark$^\ddagger$        &       \\
% Touchdown wordcount based on 9326 instructions x 108 tokens avg length.
REVERIE   & 1         &       22K\pz       &   388K    & 7K & &   \checkmark     &     \checkmark$^\ddagger$       &       \\
RxR  & 3         &       126K\pz       &   9.8M    & 16.5K & &   \checkmark        &  \checkmark\pz          &    \checkmark  \\
\midrule
\multicolumn{9}{l}{\scriptsize{$^\dagger$The number of dialogues. $^\ddagger$Grounding limited to one object per instruction.}} \\
\end{tabularx}
\caption{VLN dataset comparison. RxR is larger, multilingual, and includes dense spatiotemporal groundings (Ground) and follower demonstrations (Demos).}
\label{tab:datasets}
\end{table}

\begin{figure}
\centering 
\scriptsize
\begin{tabularx}{\linewidth}{X X}
\multicolumn{1}{c}{\small \textbf{Guide Alignment}} & \multicolumn{1}{c}{\small \textbf{Follower Alignment}} \\[1mm]
\includegraphics[width=0.49\columnwidth]{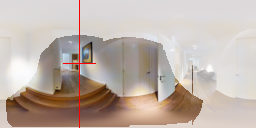} & 
\includegraphics[width=0.49\columnwidth]{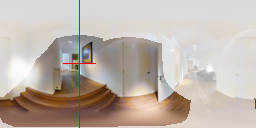} \\
Now you are standing in-front of a closed door, turn to your left, you can see two wooden steps, climb the steps and walk forward by crossing a...&
Now you are standing in-front of a closed door, turn to your left, you can see two wooden steps, climb the steps and walk forward by...\\[1mm]
\includegraphics[width=0.49\columnwidth]{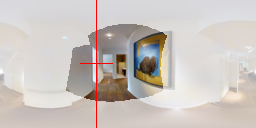} & 
\includegraphics[width=0.49\columnwidth]{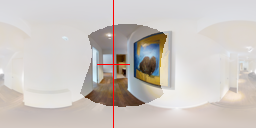} \\
...crossing a wall painting which is to your right side, you can see open door enter...&
...by crossing a wall painting which is to your right..\\[1mm]
\includegraphics[width=0.49\columnwidth]{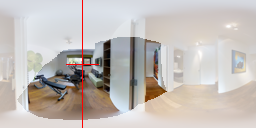} & 
\includegraphics[width=0.49\columnwidth]{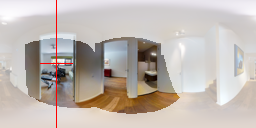} \\
...enter into it. This is a gym room, move forward, walk...&
...right side, you can see open door enter into it. This is a gym room, move forward, walk...\\[1mm]
\includegraphics[width=0.49\columnwidth]{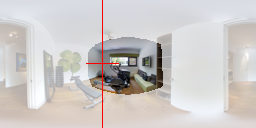} & 
\includegraphics[width=0.49\columnwidth]{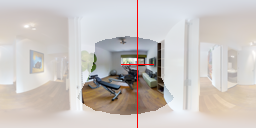} \\
...walk till the end of the room, you can see a grey...&
...walk till the end of the room, you can...\\[1mm]
\includegraphics[width=0.49\columnwidth]{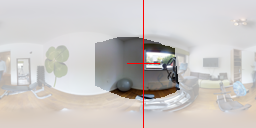} & 
\includegraphics[width=0.49\columnwidth]{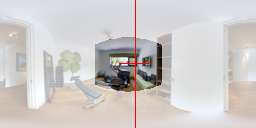} \\
...grey colored ball to the corner of the room, stand there, that's...&
...can see a grey colored ball to the corner...\\[1mm]
\includegraphics[width=0.49\columnwidth]{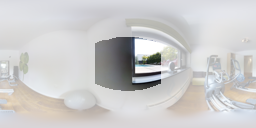} & 
\includegraphics[width=0.49\columnwidth]{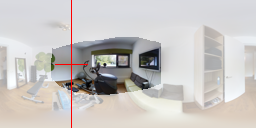} \\
...that's your end point. &
...corner of the room, stand there...\\[1mm]
 & 
\includegraphics[width=0.49\columnwidth]{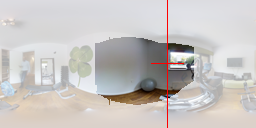} \\
 &
...that's your end point. \\[1mm]
 & 
\includegraphics[width=0.49\columnwidth]{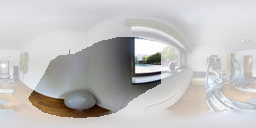} \\

\end{tabularx}
\caption{Example spatiotemporal alignment of textual instructions, visual percepts and actions for an en-US Guide and the corresponding Follower. The next selected action is indicated in red and unseen pixels in the equirectangular panoramic images are faded. The Follower takes a slightly longer path but produces similar visual-textual alignments. Best viewed enlarged. }
\label{fig:gf-traces}
\end{figure}

\textbf{Fine-Grained Grounding.}
Like R2R, RxR's instructions are collected by immersing \textit{Guide} annotators in a simulated first-person environment backed by the Matterport3D dataset \cite{Matterport3D} and asking them to describe predefined paths. RxR also enhances each instruction with dense spatiotemporal groundings. Guides speak as they move and later transcribe their audio; our annotation tool records their 3D poses and time-aligns the entire \textit{pose trace} with words in the transcription. Instructions and pose traces can thus be aligned with any Matterport data including surface reconstructions (\figref{fig:teaser}), RGB-D panoramas (\figref{fig:gf-traces}), and 2D and 3D semantic segmentations. 

\textbf{Follower Demonstrations.}
Annotators also act as \textit{Followers} who listen to a Guide's instructions and attempt to follow the path. In addition to verifying instruction quality, this allows us to collect a play-by-play account of how a human interpreted the instructions, represented as a pose trace. Guide and Follower pose traces provide dense spatiotemporal alignments between instructions, visual percepts and actions -- and both perspectives are useful for agent training.

\textbf{Path Desiderata.}
R2R paths span 4--6 edges and are the shortest paths from start to goal. \citet{thomason:naacl19} showed that agents can exploit effective priors over R2R paths, and \citet{jain2019stay} showed that R2R paths encourage goal seeking over path adherence. These matter both for generalization to new environments and fidelity to the descriptions given in the instruction---otherwise, strong performance might be achieved by agents that mostly ignore the language. RxR addresses these biases by satisfying four \textit{path desiderata}: 

\begin{enumerate}
    \item High variance in path length, such that agents cannot simply exploit a strong length prior.
    \item Paths may approach their goal indirectly, so agents cannot simply go straight to the goal.
    \item Naturalness: paths should not enter cycles or make continual direction changes that would be difficult for people to describe and follow.
    \item Uniform coverage of environment viewpoints, to maximize the diversity of references to visual landmarks and objects over all paths.
\end{enumerate}

\noindent
This increases RxR's utility for testing agents' ability to ground language. It also makes RxR a more challenging VLN dataset---but one for which human followers still achieve a 93.9\% success rate. 

\section{Two-Level Path Sampling}
\label{sec:sampling}

We satisfy desiderata 1-3 using a two-level procedure. At a high-level, each path visits a sequence of rooms; these are \textit{simple paths} with no repeated (room) vertices. Such paths are \textit{not necessarily} shortest paths. The low-level sequence is then the shortest panorama path, constrained by the room sequence. Given the set of all such paths across all houses, the fourth desiderata is satisfied by iteratively selecting the path that most improves coverage while maintaining a bias against shortest paths.

\begin{figure}
\begin{subfigure}{0.49\columnwidth}
    \includegraphics[clip,trim={.8cm .5cm .3cm 0},width=\linewidth]{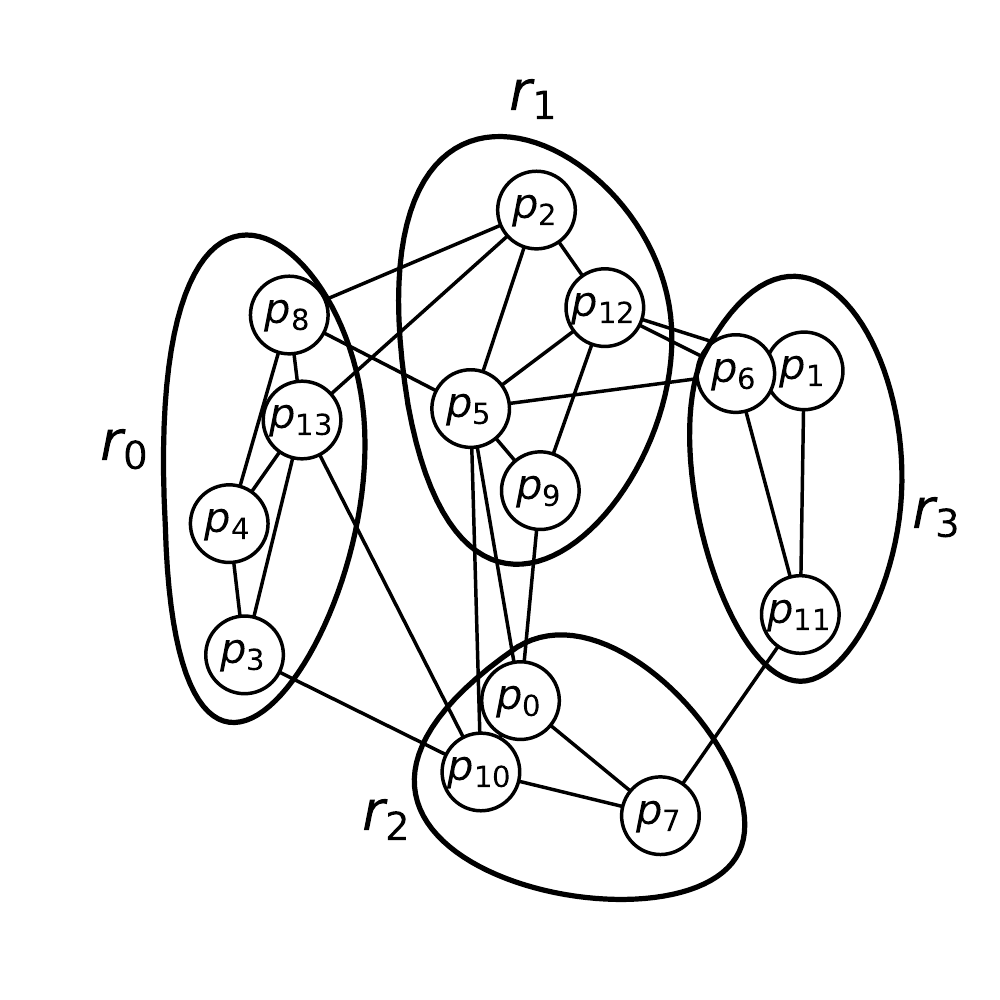}
    \caption{Panorama navigation graph with room annotations.}
    \label{fig:sampling_a}
\end{subfigure}
\hspace*{\fill}
\begin{subfigure}{0.49\columnwidth}
    \includegraphics[clip,trim={.8cm .5cm .3cm 0},width=\linewidth]{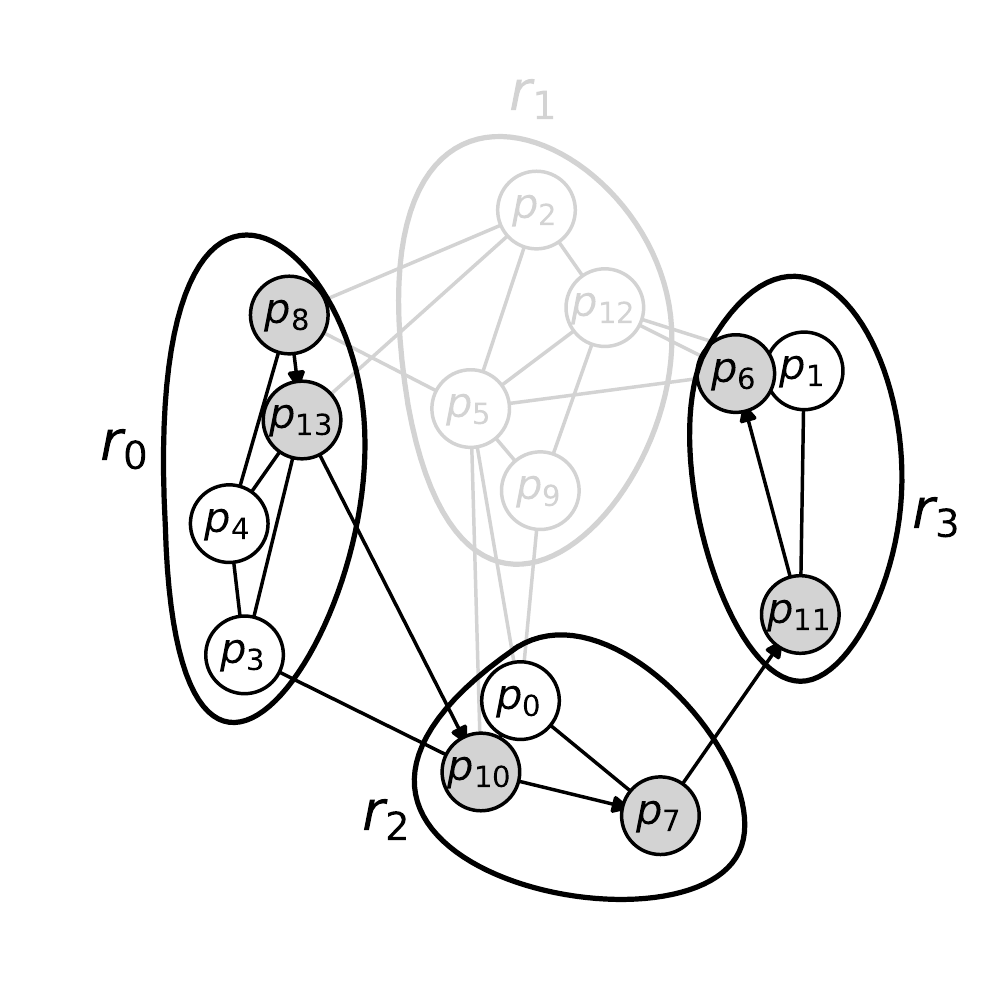}
    \caption{Shortest path from $p_8$ to $p_{6}$ on the subgraph.}
    \label{fig:sampling_b}
\end{subfigure}
\caption{Given the panorama navigation graph $P$ with room graph $R$ in \figref{fig:sampling_a}, we sample a simple room path $(r_0,r_2,r_3)$ inducing the subgraph in \figref{fig:sampling_b}. The generated panorama path is the shortest path in the subgraph linking sampled panoramas $r_8$ and $r_6$.}
\label{fig:sampling}
\end{figure}

\paragraph{Preliminaries} 
Movement in the simulator is based on a navigation graph. Vertices correspond to 360-degree panoramic images, captured at approximately 2.2m intervals throughout 90 indoor environments. Edges are navigable links between panoramas. 
\citet{Matterport3D} also partition panoramas via human-defined \textit{room} annotations.

Let \graphp be an undirected graph of interconnected panoramas, with vertices $p_i \in \mathcal{V}(\graphp)$ and edges $(p_i, p_j){\in}\mathcal{E}(\graphp)$. Let $A_R$ be a set of disjoint room annotations; each room $r_i{\in}A_R$ is a non-overlapping subset of panoramas $r_i \subseteq \mathcal{V}(\graphp)$, as shown in \figref{fig:sampling_a}. We abbreviate $(p_1, \cdots, p_m)$ as $p_{1:m}$.

We create \graphr, an undirected room graph with vertices $\mathcal{V}(\graphr) = \{\bigcup \mathcal{C}(P[r_i]) \mid r_i{\in}A_R\}$. $P[r_i]$ is the subgraph of \graphp induced by room annotation $r_i$ and $\mathcal{C}$ returns a graph's connected components. Simply put, each vertex in \graphr encompasses a subgraph of \graphp. 
An edge $(r_i, r_j) \in \mathcal{E}(\graphr)$ exists if the subgraph of \graphp induced by $\mathcal{V}(r_i) \cup \mathcal{V}(r_j)$ is connected. 

\paragraph{Path Generation}
We generate the set of all simple paths in $R$ that traverse at most 5 rooms and two building levels. Let $r_{p_i}{\in}\mathcal{V}(R)$ be the room containing panorama $p_i$. As shown in \figref{fig:sampling_b}, for each room path $r_{1:n}$, we construct a directed graph $P[r_{1:n}]$ in which an edge $(p_i, p_j)$ exists if $r_{p_i}{=}r_{p_j}$ ($p_i$ and $p_j$ are in the same room) or $(r_{p_i},  r_{p_{j}})$ is an edge in the room path. Given $P[r_{1:n}]$, we sample the start $p_1$ and goal $p_m$ uniformly from $r_1$ and $r_n$, respectively. The full panorama path $p_{1:m}$ is then the shortest path between $p_1$ and $p_m$ in $P[r_{1:n}]$. 
%Let the set of all such paths be $M$.

Room size varies greatly, so this approach produces high path length variance. It also satisfies naturalness because people tend to ground instructions at the room level (\eg \textit{Exit through the carved wooden door on the other side of the room}). We find such paths easy to describe even with as many as 20 edges. Finally, these paths can approach their goal indirectly, as exemplified in \figref{fig:sampling_b}.

\begin{figure}
\begin{center}
\includegraphics[clip,trim={0.3cm 0.3cm 0.3cm 0.3cm},width=0.99\linewidth]{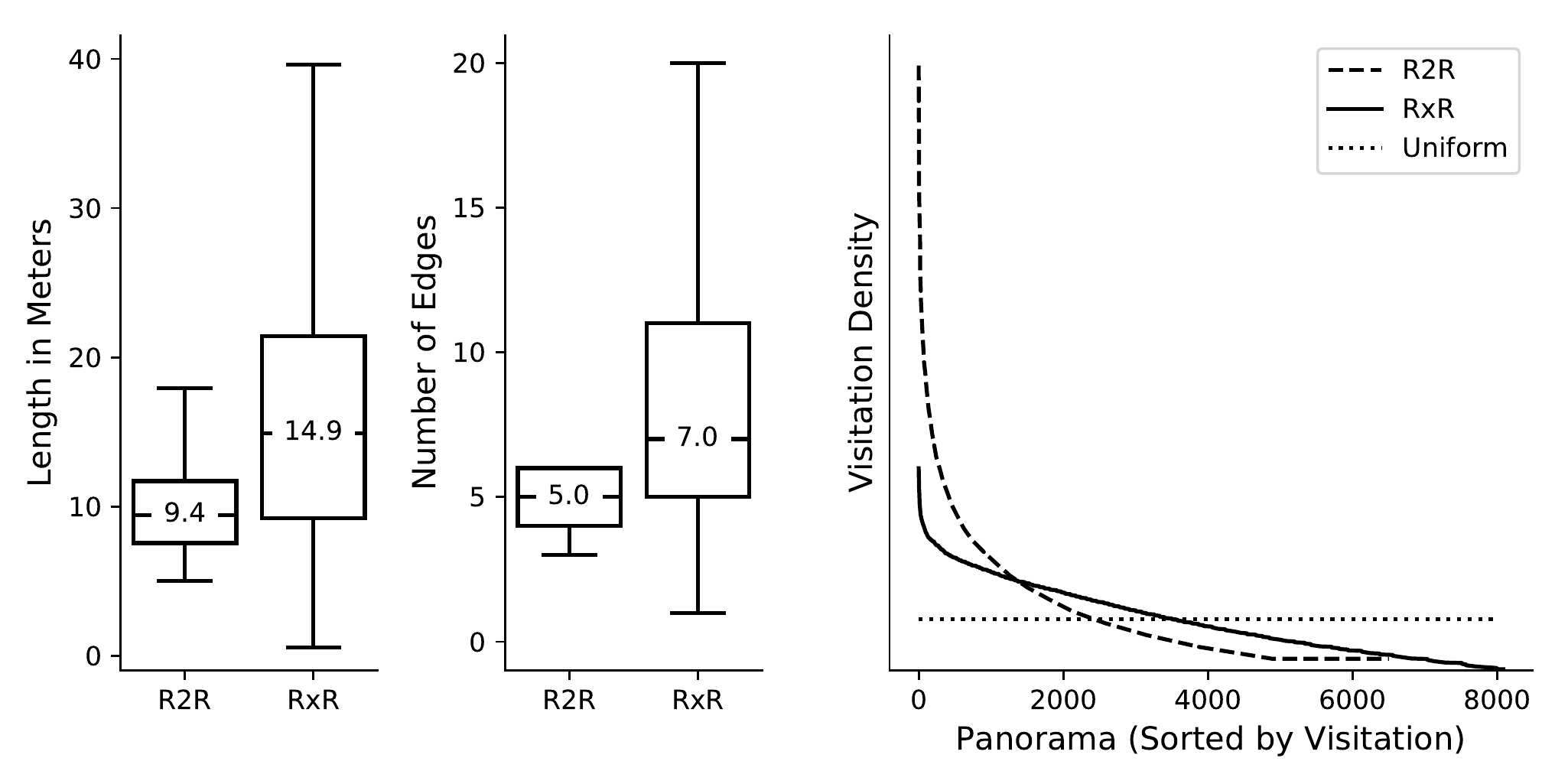}
\end{center}
\caption{RxR's paths are longer on average than R2R's, exhibiting far greater variation in length (measured in both meters and edges) while achieving more uniform panorama coverage. Comparisons shown are for train and val since the R2R test set is sequestered.}
\label{fig:paths}
\end{figure}

\paragraph{Greedy Selection for Coverage}
The final path dataset $D$ is constructed by repeatedly selecting a panorama path $p_{1:m}$ from all sampled paths (without replacement) until a desired size is reached. After selecting $k$ paths, let $\O(p_i, D_k)$ be the number of occurrences of panorama $p_i$ in the paths in $D_k$. At step $k+1$, we select the path with the minimum value for $\frac{d(p_1, p_m)}{L(p_{1:m})} + \frac{1}{m}\sum_{p_i \in p_{1:m}} \O(p_i, D_k)$, where $L$ is path length in $P$ and $d(p_1, p_m)$ is the shortest path distance between $p_1$ and $p_m$ in $P$. The first term prefers non-shortest paths while the second encourages selection of paths that cover panoramas with low coverage in $D_k$. This selection step is also subject to a maximum path length of 40m, and a maximum of 500 paths per building environment. 

\paragraph{Path Statistics}\label{par:path-stats}
In total, we sample 16522 paths, which are split: 11089 train, 1232 val-seen (train environments), 1517 val-unseen (val environments), and 2684 test, following the same environment splits as Matterport3D and R2R. Compared to R2R, RxR paths are longer, spanning 8 edges and 14.9m on average, vs. 5 edges and 9.4m in R2R. More importantly, as shown in \figref{fig:paths}, RxR paths exhibit much greater variation in length while also achieving more uniform coverage of the panoramas (and edges). Furthermore unlike R2R, 44.5\% of RxR paths are \textit{not} the shortest path from the start to the goal location. RxR paths are on average 27.4\% longer than the shortest path.

\section{Data Collection and Metrics}\label{sec:collection}

We immerse annotators in our own web-based version of the Matterport3D simulator using the panoramic images and the navigation graph. Compared to 
\citet{mattersim}, our annotation tool has additional capabilities including speech collection, virtual pose tracking, and time-alignment between transcript and pose. \figref{fig:gf-traces} gives an example instruction with accompanying Guide and Follower pose traces.
Here, we describe our collection process, analysis of the data, path evaluation metrics and simple baselines.

\label{sec:guide}

\paragraph{Guide Task}
Like R2R, our simulator has camera controls allowing continuous heading and elevation changes and movement between panoramas. Guides look around and move to explore a provided path and attempt to create an instruction others can follow. R2R's Guides create written instructions. In contrast, RxR's Guides \textit{speak} and the tool logs their entire virtual camera pose sequence. We use a 640 $\times$ 480 pixel viewing canvas and a camera vertical field of view of 75 degrees. This process is inspired by Localized Narratives \cite{PontTuset_eccv2020}, an image captioning dataset for which annotators move mouse pointers around images while talking about them.

\begin{table}
\centering
\footnotesize
\setlength\tabcolsep{1.7pt}
\begin{tabularx}{\columnwidth}{Xrr|rrr|r} &
    \textbf{en-IN} &
    \textbf{en-US} &
    \textbf{en} &
    \textbf{hi} &
    \textbf{te} &
    \textbf{Total} \\
\midrule
\textbf{Counts} & & & & & \\
Instructions & 28010 & 13992 & 42002 & 42068 & 41999 & 126069 \\
Paths & 14005 & 13992 & 14005 & 14026 & 14003 & 16522 \\
% Instructions & 28K & 14K & 42K & 42K & 42K & 126K \\
% Paths & 14K & 14K & 14K & 14K & 14K & 16.5K \\
\midrule
\textbf{Averages} & & & & & \\
Words & 87 & 129 & 101 & 76 & 56 & 78 \\
WordPieces & 104 & 159 & 123 & 143 & 184 & 150 \\
Characters & 457 & 659 & 524 & 355 & 395 & 425 \\
Audio (s) & 64 & 80 & 69 & 53 & 58 & 60 \\
Guide (s) & 431 & 509 & 457 & 451 & 465 & 458 \\
Follower (s) & 134 & 202 & 156 & 110 & 132 & 132 \\
\end{tabularx}
\caption{RxR summary statistics. Times in seconds (s).}
\label{tab:summary}
\end{table}

\setlength{\tabcolsep}{3.0pt}
\begin{table*}
\small
\begin{center}
\begin{tabularx}{\textwidth}{lccccccccccX}
%{r\hspace{10mm}cc\hspace{10mm}@{\hspace{8mm}}cc@{\hspace{8mm}}cc@{\hspace{8mm}}cc@{\hspace{8mm}}l}
%&  \multicolumn{2}{c@{\hspace{10mm}}}{\textbf{R2R} } &  \multicolumn{8}{c@{\hspace{8mm}}}{\textbf{RxR} } &  \multicolumn{1}{c}{}\\[2mm]
%\cline{2-12}
%\cline{4-11}

&  \multicolumn{2}{c}{\textbf{R2R} } &  \multicolumn{8}{c}{\textbf{RxR} } &  \multicolumn{1}{c}{}\\
\cmidrule{4-11}
&  \multicolumn{2}{c}{{en} } &  \multicolumn{2}{c}{{hi} } &  \multicolumn{2}{c}{{te} } &  \multicolumn{2}{c}{{en-IN} } &  \multicolumn{2}{c}{{en-US} } &  \multicolumn{1}{c}{}\\
\bf Phenomenon & \bf $p$ & \bf $\mu$ & \bf $p$ & \bf $\mu$  & \bf $p$ & \bf $\mu$  & \bf $p$ & \bf $\mu$  & \bf $p$ & \bf $\mu$ &  \multicolumn{1}{c}{\textbf{RxR Example (en-US)}}\\
\midrule
Reference  & 100 & 3.7 & 100 & 5.8 & 100 & 6.6 & 100 & 6.4 & 100 & 8.3 & \quad{...there is \textbf{a white chair} and \textbf{a table stand}...} \\
Coreference & \pz32 & 0.5 & \pz40 & 0.4 & \pz76 & 2.9 & \pz76 & 6.4 & \pz64 & 5.3 & \quad{...hallway with black curtains, towards \textbf{that}...} \\
Comparison & \pzz4 & 0.0 & \pzz0 & 0.0 & \pzz4 & 0.1 & \pzz4 & 0.0 & \pzz8 & 0.0 & \quad{...the large archway with the \textbf{smaller} archway in...} \\
Sequencing & \pz16 & 0.2 & \pz24 & 0.2 & \pz44 & 0.6 & \pz44 & 0.5 & \pz52 & 0.9 & \quad{...the \textbf{next} room... turn to see the \textbf{next} door...}\\
Allocentric Relation & \pz20 & 0.2 & \pz68 & 2.1 & \pz76 & 3.2 & \pz92 & 3.4 & \pz76 & 2.4 & \quad{...a window with a black folding table \textbf{under} that...} \\
Egocentric Relation & \pz80 & 1.2 & \pz96 & 2.9 & \pz80 & 2.3 & \pz64 & 2.8 & \pz60 & 2.3 & \quad{...chairs on \textbf{your right}, closet doors on \textbf{your left}.} \\
Imperative & 100 & 4.0 & 100 & 5.6 & 100 & 6.5 & 100 & 8.4 & 100 & 6.3 & \quad{\textbf{Do not} go down the stairs. Instead, \textbf{look} further...}\\
Direction & 100 & 2.8 & \pz96 & 5.8 & \pz96 & 4.9 & 100 & 7.0 & \pz96 & 6.3 & \quad{...\textbf{veer to the left} of the fireplace and you will...}\\
Temporal Condition & \pz28 & 0.4 & \pz32 & 0.4 & \pz36 & 0.7 & \pz44 & 1.0 & \pz52 &  0.8 & \quad{Move around the island \textbf{until} you come to the...} \\
State Verification & \pzz8 & 0.1 & \pz72 & 1.7 & \pz68 & 1.6 & \pz80 & 2.3 & \pz84 & 3.1 & \quad{...\textbf{you are in} the balcony area facing towards...}\\
\end{tabularx}
\end{center}
\caption{Linguistic phenomena in a manually annotated random sample of 25 paths from RxR and R2R. $p$ is the \% of sentences that contain the phenomena while $\mu$ is the average number of times they occur within each sentence.
}
\label{tab:phenomena}
\end{table*}
\setlength{\tabcolsep}{6pt}
\renewcommand{\arraystretch}{1.0}

As with Localized Narratives, RxR Guides transcribe their own recordings; this produces high quality text versions of the instructions. To align text and pose traces, we generate a time-stamped transcription using automatic speech recognition.\footnote{\href{https://cloud.google.com/speech-to-text}{https://cloud.google.com/speech-to-text}} The transcription and ASR output are aligned using dynamic time warping. The output of the Guide task is an audio file, a tokenized, timestamped, manually-transcribed instruction, and a \textit{pose trace} (a series of timestamped 6-DOF camera poses). On average, Guide task annotations  (including both steps, performed back-to-back) take 458 seconds. 

For each language (English, Hindi and Telugu) we annotate 14K paths with three instructions each. In the English dataset, each path gets one US English instruction and two Indian English instructions. Of the 14K paths per language, 12.8K paths are common across all three languages, and 1.2K paths in each language are unique (equaling 16.5K paths in total). The fact that most paths are annotated 9 times (3 per language) creates interesting opportunities to study aligned instructions across languages. Unique paths add variety and coverage.

\label{sec:follower}

\paragraph{Follower Task}
As Followers, annotators begin at the start of an unknown path and try to follow the Guide's instruction. They observe the environment and navigate in the simulator as the Guide's audio plays. They can pause, rewind and skip forward in the instruction. If they believe they have reached the the end of the path, or give up, they indicate they are done and rate the instruction's clarity and their confidence in their own navigation. On average, Follower tasks take 132 seconds. 

The Follower tasks objectively validate the quality of Guide instructions based on whether the Follower can succeed (\ie reaching within 3m of the last panorama in the path). If the Follower doesn't succeed, the Guide instruction is paired with a second Follower. If the second Follower succeeds, the first Follower annotation is discarded and replaced. If the second Follower also fails, then the path is re-enqueued to generate another Guide and Follower annotation. The most successful of the three resulting Guide-Follower pairs is selected for inclusion in RxR and the others are discarded.

In addition to validating data quality, the Follower task also trains annotators to be better Guides---following bad instructions often helps one see how to produce better instructions. Most importantly, we collect the pose trace of the Follower as they execute the instruction. This provides an alternative path with dense grounding that we can compare to the Guide's pose trace and use as an additional training signal.

\paragraph{Dataset Analysis}
\tabref{tab:summary} provides summary statistics for RxR.
The average words per instruction (using whitespace tokenization) is 78 vs R2R's 29. %(consistent with RxR's longer paths). 
US English instructions are the longest on average. We attribute this to conventions developed by each annotator pool rather than language specific properties. On average Guide tasks take much longer than Follower tasks (458 vs. 132 seconds). Most of the Guide's time is spent transcribing audio (Guide audio recordings average 60 seconds). 

Following a similar analysis as \citet{chen2019touchdown}, \tabref{tab:phenomena} gives examples and statistics for linguistic phenomena, based on manual analysis of instructions for 25 paths. All RxR subsets produce a higher rate of \textit{entity references} compared to R2R. This is consistent with the extra challenge of RxR's paths and our annotation guidance that instructions should help followers stay on the path as well as reach the goal. Doing so requires more extensive use of objects in the environment. RxR's higher rate of both \textit{coreference} and \textit{sequencing} indicates that its instructions have greater discourse coherence and connection than R2R's. RxR also includes a far higher proportion of \textit{allocentric relations} and \textit{state verification} compared to R2R, and matches Touchdown (navigation instructions). Hindi contains less coreference, sequencing, and temporal conditions than the other languages. That said, it is not clear how much the differences \textit{within} RxR exhibited in \tabref{tab:phenomena} can be attributed to language, dialect, annotator pools, or other factors.

\begin{figure}
\begin{center}
\includegraphics[clip,trim={0.2cm 0 0.3cm 0},width=\linewidth]{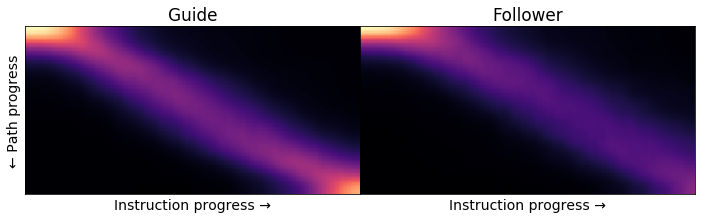}
\includegraphics[clip,trim={0.3cm 0 0.3cm 0},width=\linewidth]{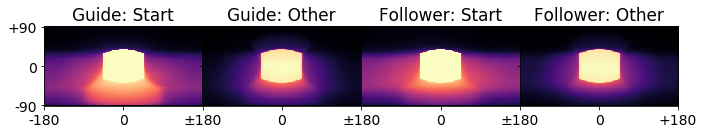}
\end{center}
\caption{Top: Instruction and path progress alignment for Guides and Followers. Bottom: Equirectangular heatmap of Guide and Follower camera poses, centered on their initial perspective at each viewpoint. }
\label{fig:gaze}
\end{figure}

\figref{fig:gaze} (top) illustrates the close alignment between instruction progress (measured in words) and path progress (measured in steps). \figref{fig:gaze} (bottom) indicates that both Guide and Tourist annotators orient themselves by looking around at the first panoramic viewpoint, after which they maintain a narrower focus. On average, Guides / Tourists observe 43\% / 44\% of the available spherical visual signal at the first viewpoint, and 27\% / 28\% at subsequent viewpoints. These findings stand in contrast to standard VLN agents that routinely consume the entire panoramic image and attend over the entire instruction sequence at each step. Inputs that the Guide / Tourist have not observed cannot influence their utterances / actions, so pose traces offer rich opportunities for agent supervision.

\begin{table}
\resizebox{8cm}{!}{
\centering 
\begin{tabular}{lcccccc}
& \bf PL & \bf NE$\downarrow$ & \bf SR$\uparrow$ & \bf SPL$\uparrow$ & \bf SDTW$\uparrow$ & \bf NDTW$\uparrow$ \\
\midrule
\multicolumn{7}{l}{1. Random walk} \\[1mm]
  R2R & 
10.4 & \pz9.5 & \pz5.1 & \pz3.6 & \pz3.8 & 27.6
\\
RxR & 
16.8 & 12.4 & \pz8.8 & \pz2.5 & \pz3.8 & 18.2
\\
\midrule
\multicolumn{7}{l}{2. Random heading then go straight} \\[1mm]
  R2R & 
\pz9.7 & \pz9.9 & \pz8.2 & \pz7.2 & \pz6.6 & 28.3
\\
 RxR & 
15.1 & 13.5 & \pz8.0 & \pz3.4 & \pz3.9 & 16.3
\\
\midrule
\multicolumn{7}{l}{3. Given correct first step then go straight} \\[1mm]
R2R & 
\pz9.5 & \pz6.2 & 27.2 & 25.7 & 23.6 & 52.6
\\
RxR & 
15.3 & 11.4 & 13.7 & \pz7.5 & \pz8.3 & 25.9
\\
\end{tabular}
}
\caption{Simple baselines on val-unseen paths. RxR proves more difficult than R2R overall, and less amenable to agents that tend to go straight (baselines 2 and 3). Note: Baseline 3 partly exploits the gold path.
}
\label{tab:baselines}
\end{table}

\label{sec:eval}

\paragraph{Evaluation}
We use the following standard evaluation metrics (with arrows indicating improvement): 
\textit{Path Length} (\textbf{PL}), 
\textit{Navigation Error} (\textbf{NE~$\downarrow$})
\textit{Success Rate} (\textbf{SR~$\uparrow$}),
\textit{Success weighted by inverse Path Length} (\textbf{SPL~$\uparrow$}),
\textit{Normalized Dynamic Time Warping} (\textbf{NDTW~$\uparrow$}), and 
\textit{Success weighted by normalized Dynamic Time Warping} (\textbf{SDTW~$\uparrow$}). See \citet{evaluation2018} and 
\citet{magalhaes2019effective} for discussion of VLN metrics. Since RxR was designed to include paths that approach their goal indirectly, we focus primarily on \textbf{NDTW} and \textbf{SDTW} which explicitly capture path adherence. %\todo{Recommendation for primary metric on RxR and justification}. 
See \tabref{tab:baselines} for a comparison of the performance of several simple baselines on R2R and RxR. Each simple baseline requires a stopping criteria; we choose to stop after $N$ steps where $N$ is the average number of steps in the train set paths (5 in R2R and 8 in RxR). Consistent with our motivation to reduce biases in paths, these simple baselines show that going straight is far less effective in RxR than R2R.

\section{Experiments}

\paragraph{Agent} We use a model architecture similar to that of the Reinforced Cross-Modal Matching (RCM) agent \cite{wang2018reinforced}, consisting of an instruction encoder and a sequential LSTM \cite{Hochreiter1997} decoder that computes a distribution over actions at each step. However, since RxR instructions are much longer than R2R, we replace the bidirectional LSTM instruction encoder with a more parallelizable CNN encoder. In preliminary experiments on R2R we find that encoding word embeddings via successive 1D convolutions with rectified linear (ReLU) activations and residual connections \cite{he2015deep} is equally effective and more time and space efficient. We denote the output of the instruction encoder by $x \in \mathbb{R}^{l\times d}$ where $l$ is the instruction length and $d$ is the feature dimension. In both monolingual and multilingual experiments we use features extracted from a pre-trained multilingual BERT model \cite{devlin2018bert} for the word embeddings.

At each time step $t$, the agent receives a panoptic encoding of its viewpoint $v_t \in \mathbb{R}^{k\times d}$ (where $k=36$ is the number of $30^\circ$ intervals that span the panorama) along with a visual encoding of navigable directions $a_t \in \mathbb{R}^{n\times d}$ (where $n$ is the number of navigable directions). Each feature of dimension $d$ is a pre-trained CNN feature concatenated with an angle encoding \cite{fried2018speaker}. The LSTM decoder computes an updated hidden state $h_t$ by conditioning on the previous selected action in $a_{t-1}$ and attending over the panoptic encoding $v_t$ and the instruction $x$ using dot-product attention \cite{luong2015effective}. The distribution over next actions is computed via a similarity ranking $h_t \cdot a_{t,i}$ between hidden state $h_t$ and each direction encoding in $a_t$.

For the image features we use an EfficientNet-B4 CNN \cite{tan2019efficientnet}. Following \citet{crisscross_2020}, we pretrain the CNN in an image-text dual encoder setting using the Conceptual Captions dataset \cite{sharma2018conceptual}. In preliminary experiments, we found that pretraining the CNN in this way gave noticeable improvements over the same CNN pretrained for image classification on ImageNet \cite{ILSVRC15}.

\begin{table*}
    \setlength{\tabcolsep}{0.3em}
	\footnotesize
	\begin{center}
    \begin{tabularx}{\linewidth}{lXccccccccccccccccccccc}	
     & & \multicolumn{3}{c}{Setting} & & Training &  & \multicolumn{3}{c}{\bf NE $\downarrow$} &  & \multicolumn{3}{c}{\bf SR $\uparrow$} &  & \multicolumn{3}{c}{\bf SDTW $\uparrow$} &  & \multicolumn{3}{c}{\bf NDTW $\uparrow$} \\
     \cmidrule{3-5} \cmidrule{9-11} \cmidrule{13-15} \cmidrule{17-19} \cmidrule{21-23}
     Exp. & Method & G & F & X & &  Pairs (K) &  & en & hi & te   &  & en & hi & te   &  & en & hi & te  &  & en & hi & te    \\
  	\midrule
    (1) & Mono & \checkmark & & & & 42 & & 10.1 & \pz9.7 & \pz9.4 && 25.6 & 24.8 & 28.0 && 20.3 & 19.7 & 22.7 && 41.3 & 38.8 & 43.7 \\
    (2) & Mono & & \checkmark & & & 42 & & 10.3 & \textbf{\pz9.2} & \pz9.5 && 23.9 & 28.0 & 27.0 && 18.5 & 22.7 & 22.0 && 37.0 & \textbf{45.9} & 43.9 \\
    (3) & Mono & \checkmark & \checkmark & & & 84 & & \textbf{\pz9.8} & \textbf{\pz9.2} & \textbf{\pz9.1} && \textbf{26.1} & \textbf{29.6} & \textbf{29.8} && \textbf{21.0} & \textbf{24.0} & \textbf{24.2} && \textbf{42.4} & 45.5 & \textbf{45.6} \\
    \midrule
    (4) & Multi & \checkmark & \checkmark & & & 252 & & \textbf{11.0} & 10.9 & 11.0 && \textbf{22.2} & \textbf{23.0} & 23.1 && \textbf{17.8} & \textbf{18.3} & \textbf{18.4} && \textbf{38.6} & 39.2 & 38.8 \\
    (5) & Multi & \checkmark & \checkmark & \checkmark & & 504 && 11.5 & 11.4 & 11.4 && 20.0 & 18.7 & 20.3 && 15.9 & 14.9 & 16.1 && 36.3 & 36.0 & 36.7 \\
    (6) & Multi*  &  \checkmark   & \checkmark & &       & 252  &  &    \textbf{11.0} & \textbf{10.7} & \textbf{10.7} & & 21.9 & 22.6 & \textbf{23.2} & & 17.5 & 18.1 & \textbf{18.4} & & \textbf{38.6} & \textbf{39.9} & \textbf{39.7} \\
    %\midrule
    (H) & Human & & & & & - & & 1.32 & 0.59 & 0.79 & & 90.4 & 96.8 & 94.7 & & 74.3 & 80.6 & 76.5 & & 77.7 & 82.2 & 79.2 \\
 	\midrule
 	\multicolumn{23}{l}{\footnotesize{Settings -- G: instruction paired with Guide paths, F: instructions paired with Follower paths, X: cross-translated instructions.}}
     \end{tabularx}
    \end{center}
	\caption{RxR val-unseen: Monolingual vs. multilingual results. Training with both Guide and Follower paths benefits all languages (exp. 3 vs. 1 and 2), monolingual outperforms multilingual (exp. 3 vs. 4), training with cross-translations hurts performance (exp. 5 vs. 4), and training with visual attention supervision gives mixed results (Multi* in exp. 6 vs 4).}
	\label{tab:rxr-results}
\end{table*}

\begin{table*}
    \setlength{\tabcolsep}{0.3em}
	\footnotesize
	\begin{center}
    \begin{tabularx}{\linewidth}{Xcccccccccccccccccccccc}	
     & \multicolumn{2}{c}{Train Data} & & \multicolumn{4}{c}{\bf SR $\uparrow$} &  & \multicolumn{4}{c}{\bf SPL $\uparrow$} &  & \multicolumn{4}{c}{\bf SDTW $\uparrow$} &  & \multicolumn{4}{c}{\bf NDTW $\uparrow$}  \\
     \cmidrule{2-3} \cmidrule{5-8} \cmidrule{10-13} \cmidrule{15-18} \cmidrule{20-23}
     Exp. & R2R & RxR & &  R2R & en & hi & te &  & R2R & en & hi & te  &  & R2R & en & hi & te  &  & R2R & en & hi & te   \\
  	\midrule
    (7) & \checkmark & & & 36.5 & 14.5 & \pz9.6 & \pz9.7 &  & 31.7 & 11.2 & \pz7.5 & \pz7.4 & & 29.5 & \pz9.8 & \pz6.3 & \pz6.1 & & 48.1 & 29.0 & 25.4 & 25.2 \\
    (4) & & \checkmark  & & 19.2 & 22.2 & 23.0 & \textbf{23.1} & & 17.7 & 19.8 & 20.7 & \textbf{20.7} & & 16.0 & 17.8 & 18.3 & \textbf{18.4} & & 43.2 & 38.6 & 39.2 & \textbf{38.8} \\
    (8) & \checkmark & \checkmark & & \textbf{37.8} & \textbf{22.5} & \textbf{23.6} & \textbf{23.1} & & \textbf{34.3} & \textbf{20.1} & \textbf{21.0} & 20.5 & & \textbf{32.0} & \textbf{18.3} & \textbf{19.2} & \textbf{18.4} & & \textbf{52.3} & \textbf{38.8} & \textbf{39.4} & 38.4 \\
 	\midrule
     \end{tabularx}
    \end{center}
	\caption{Multitask and transfer learning results on RxR and R2R val-unseen. A multitask model (exp. 8) performs best on both datasets, but domain differences thwart simple transfer learning (\ie train on X, evaluate on Y). }
	\label{tab:multitask-results}
\end{table*}

\paragraph{Grounding Supervision}
To incorporate spatiotemporal groundings into agent training, for each Guide path (G-path) and Follower path (F-path) we convert the corresponding pose trace into: (1) a sequence of text masks $b_t \in \{0, 1\}^l$ indicating which words in instruction $x$ the Guide spoke / Follower heard \textit{at or prior to} step $t$, and (2) a sequence of visual masks $M_t \in \{0, 1\}^{h \times w}$ indicating which pixels were observed in the panoramic image at $t$ (like \figref{fig:gaze} bottom). We then project and max-pool $M_t$ to a vector mask $m_t \in \{0, 1\}^{k}$ aligning to the agent's visual input features $v_t$. Zeros in $b_t$ and $m_t$ indicate irrelevant textual and visual inputs that were not observed by the annotators, and are therefore not related to their utterances and actions. 

To help prevent the agent from overfitting to superficial correlations in the training data, we use $b_t$ and $m_t$ to supervise the normalized textual and visual attention weights in the model. Specifically, during training whenever the agent is on the gold path we apply a cross-entropy loss to the visual attention weights given by $\mathcal{L}(z, m_t) = \log \sum_{i=1}^k \exp(z_i) - \log \sum_{i=1}^k m_{t,i} \exp(z_i)$, where $z$ is the vector of unnormalized logits determining attention weights via a softmax. This loss forces the attention weights on irrelevant input features towards zero. The textual version is analogous.

\paragraph{Implementation Details}
Agents are implemented in VALAN \cite{lansing2019valan}, a distributed reinforcement learning framework designed for VLN.
We use a mix of supervised learning and policy gradients. Each minibatch is constructed from 50\% behavioural cloning roll-outs (following the gold paths while minimizing cross-entropy loss), and 50\% policy gradient rollouts with reward (following paths sampled from the agent's policy). As in \citet{magalhaes2019effective}, the reward at each step is the incremental difference in NDTW, plus a linear function of navigation error after stopping. All agents are trained with Adam \cite{kingma2014adam} to convergence (100K iterations with batch size of 32 and initial learning rate of 1e-4). 

\paragraph{Monolingual Results}
\tabref{tab:rxr-results} provides results on the val-unseen split for several training settings, as well as human performance from Follower annotations. We report en-US and en-IN results together as en. Experiments 1--3 compare agents trained (1) only on G-paths, (2) only on F-paths, and (3) on both.  In contrast to algorithmically generated G-paths, each F-path reflects a grounded human interpretation of an instruction, which may deviate from the G-path because multiple correct interpretations are possible (e.g., \figref{fig:gf-traces}). For training, we do not differentiate F-paths from G-paths, and each instruction-path pair is treated as an independent example. Experiment (3) shows that including both G- and F-paths in training benefits every metric. Given the overall positive impact of F-paths, we use both path types in our further experiments.

\paragraph{Multilinguality}
For experiment (4) in \tabref{tab:rxr-results}, we train a single multilingual agent on all three languages simultaneously. While the multilingual agent sees substantially more instructions than each monolingual agent, performance is worse across all metrics. This is consistent with results in multilingual machine translation (MT) and automatic speech recognition (ASR) where adding more languages can also lead to degradation for high-resource languages \cite{aharoni2019massively,pratap2020massively}. Experiment (5) takes this one step further by obtaining translations from every instruction into the two other languages (e.g., en $\rightarrow$ hi, te) using a MT service.\footnote{\url{https://cloud.google.com/translate}\\These translations are included in the RxR data release.} Including these translations hurts performance for all languages. The fact that most G-paths are shared across languages may limit the value of automatic cross-translations. Notwithstanding the higher performance of the monolingual approaches, in the remaining experiments we focus on multilingual agents for greater scalability. 

\begin{table*}
    \setlength{\tabcolsep}{0.35em}
	\footnotesize
	\begin{center}
    \begin{tabularx}{\linewidth}{lXcccccccccccccccccc}	
     & & \multicolumn{2}{c}{Input Modalities} &  & \multicolumn{3}{c}{\bf NE $\downarrow$} &  & \multicolumn{3}{c}{\bf SR $\uparrow$} &  & \multicolumn{3}{c}{\bf SDTW $\uparrow$} &  & \multicolumn{3}{c}{\bf NDTW $\uparrow$} \\
     \cmidrule{3-4} \cmidrule{6-8} \cmidrule{10-12} \cmidrule{14-16} \cmidrule{18-20}
     Exp. & Method & Vision &  Language &  & en & hi & te   &  & en & hi & te   &  & en & hi & te  &  & en & hi & te    \\
  	\midrule
    (4) & Multi & \checkmark & \checkmark & & \textbf{11.0} & \textbf{10.9} & \textbf{11.0} && \textbf{22.2} & \textbf{23.0} & \textbf{23.1} && \textbf{17.8} & \textbf{18.3} & \textbf{18.4} && \textbf{38.6} & \textbf{39.2} & \textbf{38.8} \\
    (9) & Multi & & \checkmark & & 12.3 & 11.9 & 12.0 && 16.0 & 18.0 & 16.9 && 12.3 & 14.2 & 13.3 && 30.9 & 33.1 & 32.8 \\
    (10) & Multi & \checkmark &  & & 15.7 & 15.7 & 15.7 & & \pz7.8 & \pz7.8 & \pz7.8 & & \pz4.3 & \pz4.3 & \pz4.3 & & 16.5 & 16.5 & 16.5 \\
 	\midrule
    \end{tabularx}
    \end{center}
	\caption{Language-only and vision-only model ablations on RxR val-unseen. The language-only agent is much better than random, but both modalities are required for best performance. }
	\label{tab:ablation-results}
\end{table*}

\begin{table*}
    \setlength{\tabcolsep}{0.27em}
	\footnotesize
	\begin{center}
    \begin{tabularx}{\linewidth}{Xlcccccccccccccccccccc}	
     & & & \multicolumn{4}{c}{\bf NE $\downarrow$} &  & \multicolumn{4}{c}{\bf SR $\uparrow$} &  & \multicolumn{4}{c}{\bf SDTW $\uparrow$} &  & \multicolumn{4}{c}{\bf NDTW $\uparrow$}  \\
     \cmidrule{4-7} \cmidrule{9-12} \cmidrule{14-17} \cmidrule{19-22}
     Split & Method & &  en & hi & te & avg &  & en & hi & te & avg  &  & en & hi & te & avg  &  & en & hi & te & avg   \\
  	\midrule
        Val-Seen  
        & Mono & &  \pz9.5 & \pz9.2 & \pz9.3 & \pz9.3 && 28.6 & 29.5 & 28.3 & 28.8 && 23.2 & 24.6 & 23.7 & 23.8 && 45.4 & 47.9 & 47.1 & 46.8 \\
        & Multi & &  11.0 & 10.4 & 10.6 & 10.7 && 23.9 & 26.7 & 25.1 & 25.2 && 19.6 & 21.9 & 20.5 & 20.7 && 41.2 & 43.4 & 42.0 & 42.2  \\[1mm]
        %\midrule
        Val-Unseen 
        & Mono & &  \pz9.8 & \pz9.2 & \pz9.1 & \pz9.4 && 26.1 & 29.6 & 29.8 & 28.5 && 21.0 & 24.0 & 24.2 & 23.1 && 42.4 & 45.5 & 45.6 & 44.5 \\
        & Multi & &  11.0 & 10.9 & 11.0 & 10.9 && 22.2 & 23.0 & 23.1 & 22.8 && 17.8 & 18.3 & 18.4 & 18.2 && 38.6 & 39.2 & 38.8 & 38.9  \\[1mm]
        %\midrule
        Test-Std 
         & Mono & & 11.0 & 10.5 & 10.5 & 10.6 && 25.3 & 26.1 & 26.2 & 25.9 && 20.5 & 21.0 & 21.5 & 21.0 && 40.3 & 41.9 & 42.4 & 41.5  \\
         & Multi & &  12.0 & 11.8 & 11.8 & 11.9 && 20.8 & 21.4 & 21.6 & 21.3 && 16.8 & 17.3 & 17.3 & 17.1 && 36.7 & 37.6 & 37.4 & 37.2   \\
         & Random & &     14.1 & 14.1 &  14.1 &  14.1   &  &   \pz7.5 & \pz7.5 & \pz7.5  &  \pz7.5   &  &  \pz3.1 & \pz3.1 & \pz3.1  &  \pz3.1   &  &  15.4 & 15.4 &  15.4   &  15.4   \\
         & Human & &    \pz1.4 & \pz0.6 & \pz0.7 & \pz0.9 && 90.2 & 96.7 & 94.9 & 93.9 && 73.6 & 80.5 & 76.6 & 76.9 && 77.2 & 82.0 & 79.2 & 79.5 \\ 
 	\midrule
    \end{tabularx}
    \end{center}
	\caption{RxR test set results, based on the monolingual agents (3) and the multilingual agent (4).}
	\label{tab:test-results}
\end{table*}

\paragraph{Spatiotemporal Grounding Supervision}
\tabref{tab:rxr-results} experiment (6) incorporates a loss for spatiotemporal grounding over visual attention which gives mixed results on val-unseen (better on NDTW, NE and worse on success-based metrics) compared to (4). Applying the same approach to textual attention did not improve performance. However, we stress that this is only a preliminary investigation. Using human demonstrations to supervise visual groundings is an active area of research \cite{wu2019self,selvaraju2019taking}. As one of the first large-scale spatially-temporally aligned language datasets, RxR offers new opportunities to extend this work from images to environments.

\paragraph{Multitask and Transfer Learning}
\tabref{tab:multitask-results} reports the performance of the multilingual agent under multitask and transfer learning settings. For simplicity, the R2R model (exp. 7) is trained without data augmentation from model-generated instructions \cite{fried2018speaker,backtranslate2019} and with hyperparameters tuned for RxR. Under these settings, the multitask model (exp. 8) performs best on both datasets. However, transfer learning performance (RxR~$\rightarrow$~R2R and vice-versa) is much weaker than the in-domain results. Although RxR and R2R share the same underlying environments, we note that RxR~$\rightarrow$~R2R cannot exploit R2R's path bias, and for R2R~$\rightarrow$~RxR, the much longer paths and richer language are out-of-domain.

\paragraph{Unimodal Ablations}
\tabref{tab:ablation-results} reports the performance of the multilingual agent under settings in which we ablate either the vision or the language inputs during both training and evaluation, as advocated by \citet{thomason:naacl19}. The multimodal agent (4) outperforms both the language-only agent (9) and the vision-only agent (10), indicating that both modalities contribute to performance. The language-only agent performs better than the vision-only agent. This is likely because even without vision, parts of the instructions such as `turn left` and `go upstairs` still have meaning in the context of the navigation graph. In contrast, the vision-only model has no access to the instructions, without which the paths are highly random.

\paragraph{Test Set}
RxR includes a heldout test set, which we divide into two splits: test-standard and test-challenge. These splits will remain sequestered to support a public leaderboard and a challenge so the community can track progress and evaluate agents fairly. \tabref{tab:test-results} provides test-standard performance of the mono and multilingual agents using Guide and Follower paths, along with random and human Follower scores. While the learned agent is clearly much better than a random agent, there is a great deal of headroom to reach human performance. 

\section{Conclusion}

RxR represents a significant evolution in the scale, scope and possibilities for research on embodied language agents in simulated, photo-realistic 3D environments. RxR's paths better ensure that language itself will play a fundamental role in better agents. Evaluating on three typologically diverse languages will help the community avoid overfitting to a particular language and dataset.

We have only begun to explore the possibilities opened up by pose traces. Whereas others have retro-actively refined R2R's annotations to get alignments between sub-instructions and panorama sequences \cite{hong2020subinstruction}, RxR provides \textit{word-level} alignments to \textit{specific pixels} in panoramas. This is obtained as a by-product of significant work on the annotation tooling itself and designing the process to be more natural for Guides. Finally, every instruction is accompanied by a Follower demonstration, including a perspective camera pose trace that shows a play-by-play account of how a human interpreted the instructions given their position and progress through the path.  We have shown that these can help with agent training, but they also open up new possibilities for studying grounded language pragmatics in the VLN setting, and for training VLN agents with perspective cameras -- either in the graph-based simulator or by lifting RxR into a continuous simulator \cite{krantz_vlnce_2020}.

\section*{Acknowledgments}

We thank Sneha Kudugunta for analyzing the Telugu annotations, and the Google Data Compute team, especially Igor Karpov, Ashwin Kakarla and Christina Liu, for their tooling and annotation support for this project, Austin Waters and Su Wang for help with image features, and Daphne Luong for executive support for the data collection.

\bibliography{paper}
\bibliographystyle{acl_natbib}

\clearpage
\appendix

\section{Supplementary Material}
\label{sec:supplementary}

\paragraph{Annotators}
In total, 247 annotators contributed to RxR, with 97 based in the USA and the remainder based in India and contributing to the Indian English, Hindi and Telugu annotations. The annotators were paid hourly wages that are competitive for their locale. They have standard rights as contractors. They were fluent in the language they were tasked with.

We ensure that a Guide does not annotate the same path twice. As Followers, annotators do not follow their own Guide instructions. Furthermore, we have provided annotators multiple forms of feedback as they complete tasks. After a round of pilot instructions were collected, we provided detailed analysis of common patterns that produced poor instructions and clear guidelines for producing better instructions. Annotators provided UI suggestions and interesting corner cases to us that allowed us to refine the simulator and annotation process before kicking off the full annotation process. Throughout the process, annotators have had access to a dashboard that shows them their success rate as both Guide and Follower. We indicated that their success as Guide and a Follower should be above 80\%. Any annotator whose success is lower is either given further training or is taken off the task.

Unfortunately, we cannot release the audio instructions yet due to the impact of COVID-19: our annotators had to complete the tasks from home, so we need to review all recordings for safety and privacy. We hope to include the audio in a future release.

\begin{figure*}[t]
\begin{center}
\begin{tabularx}{\textwidth}{XX}
    \fbox{\includegraphics[width=\columnwidth]{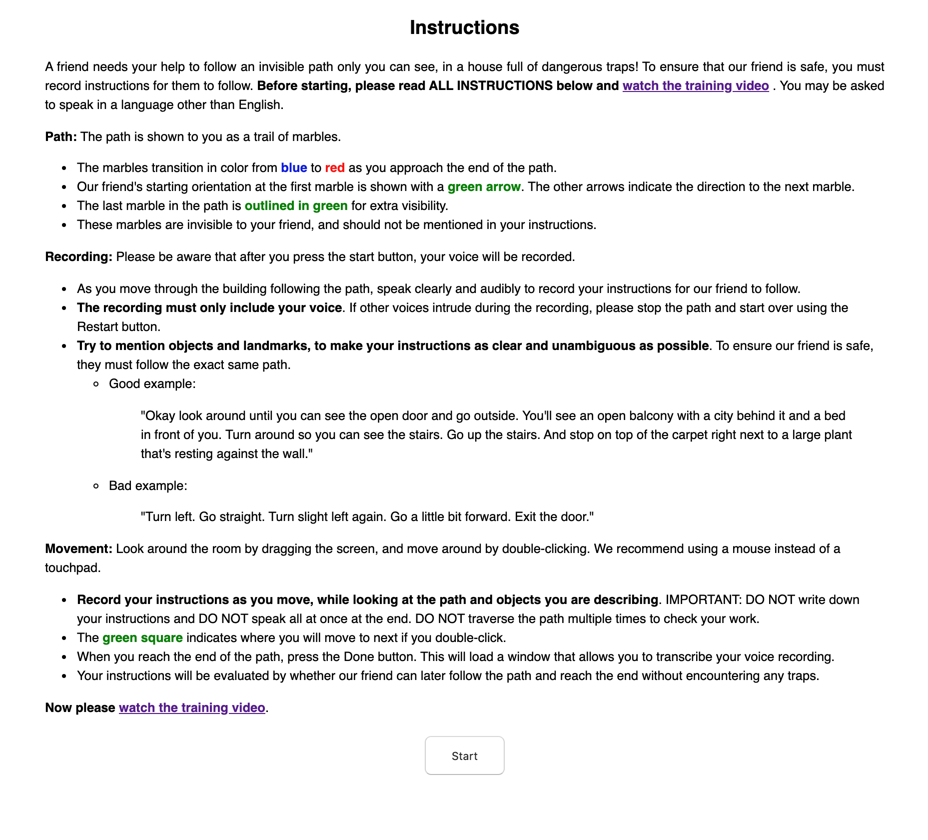}} & \fbox{\includegraphics[width=\columnwidth]{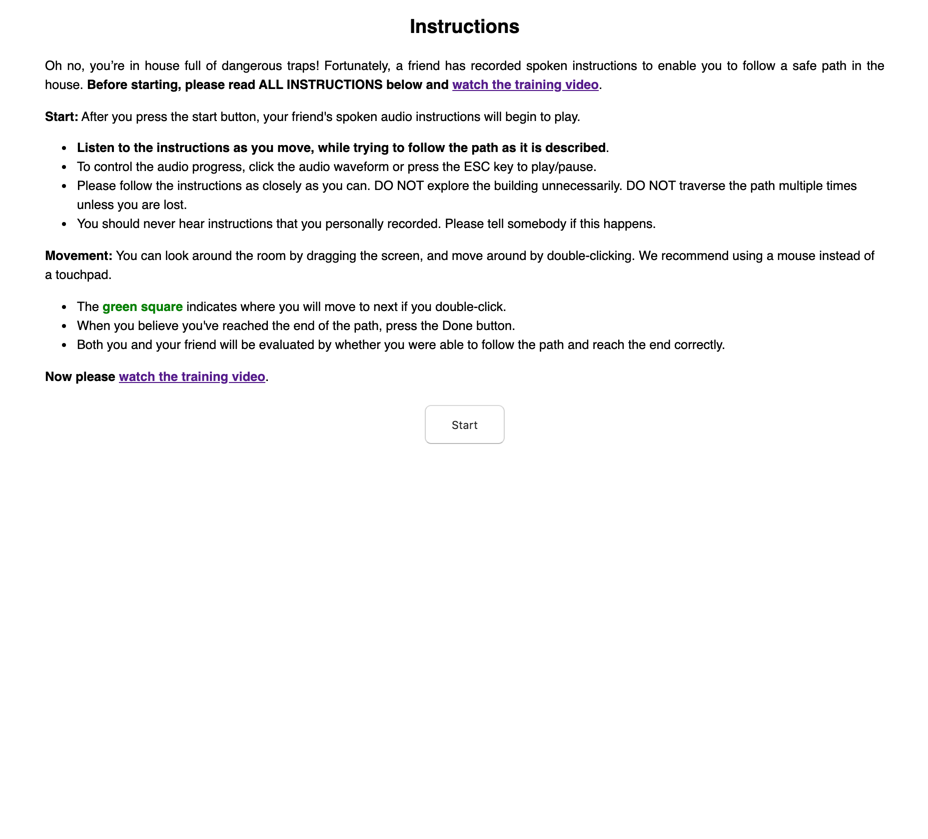}} \\
    \small (a) Guide Worker Instructions (49.1 seconds) & \small (b)  Follower Worker Instructions (38.5 seconds) \\[6pt]
    \fbox{\includegraphics[width=\columnwidth]{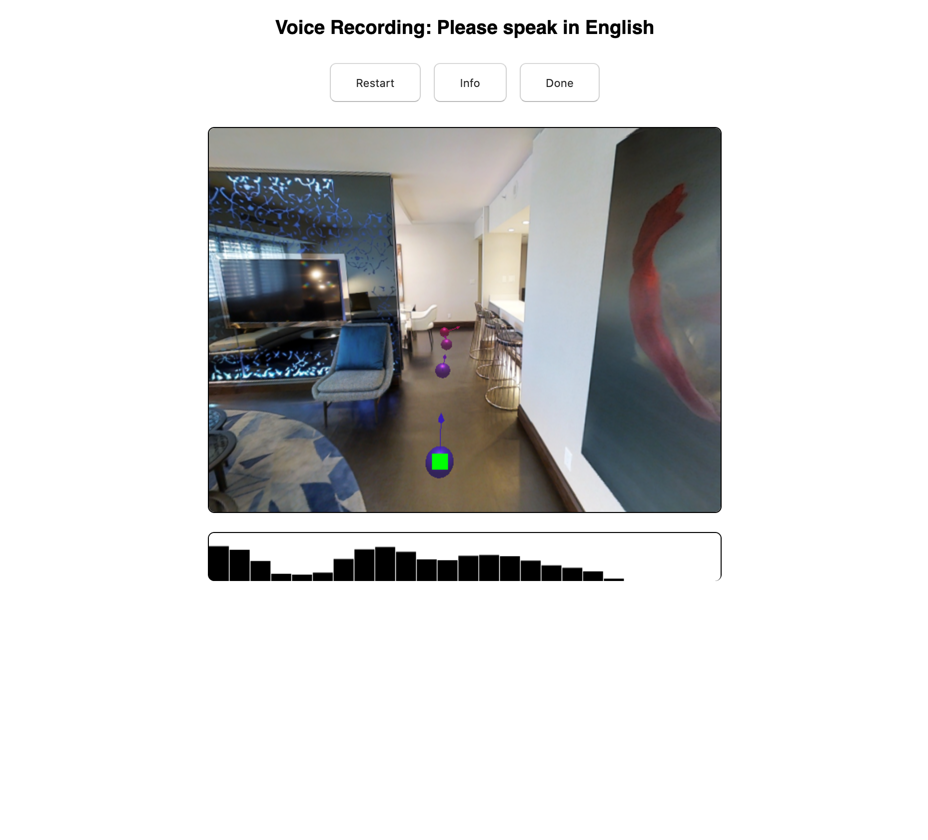}} & \fbox{\includegraphics[width=\columnwidth]{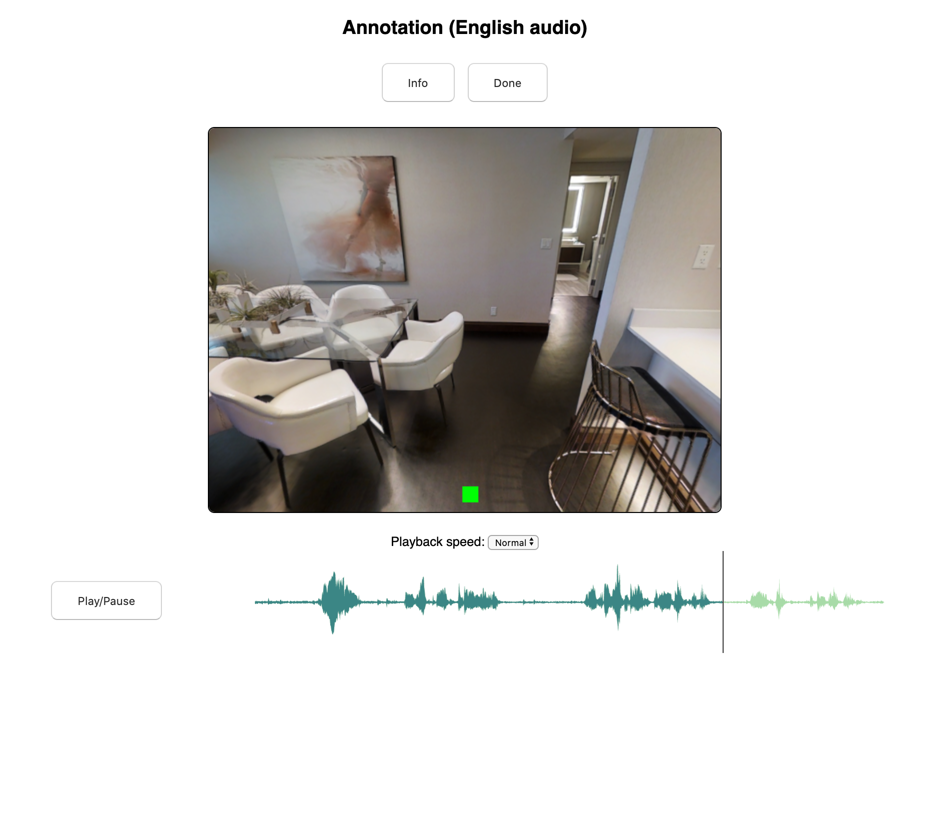}} \\
    \small (c) Guide Annotation (64.4 seconds) & \small (d) Follower Verification (89.8 seconds) \\[6pt]
    \fbox{\includegraphics[width=\columnwidth]{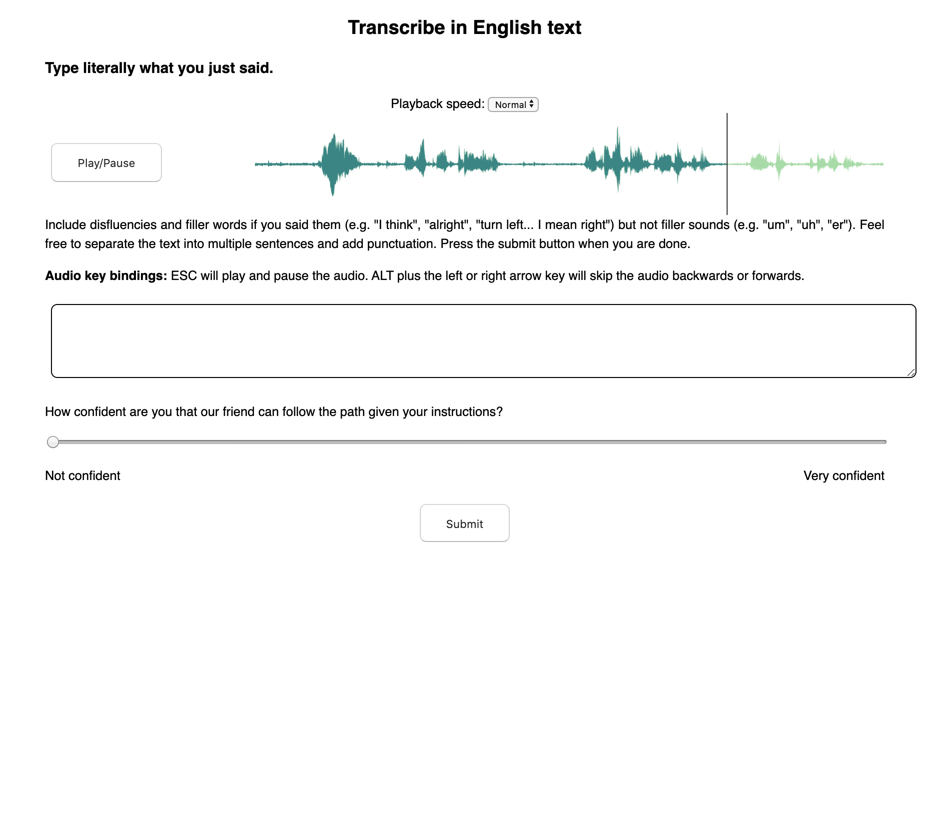}} & \fbox{\includegraphics[width=\columnwidth]{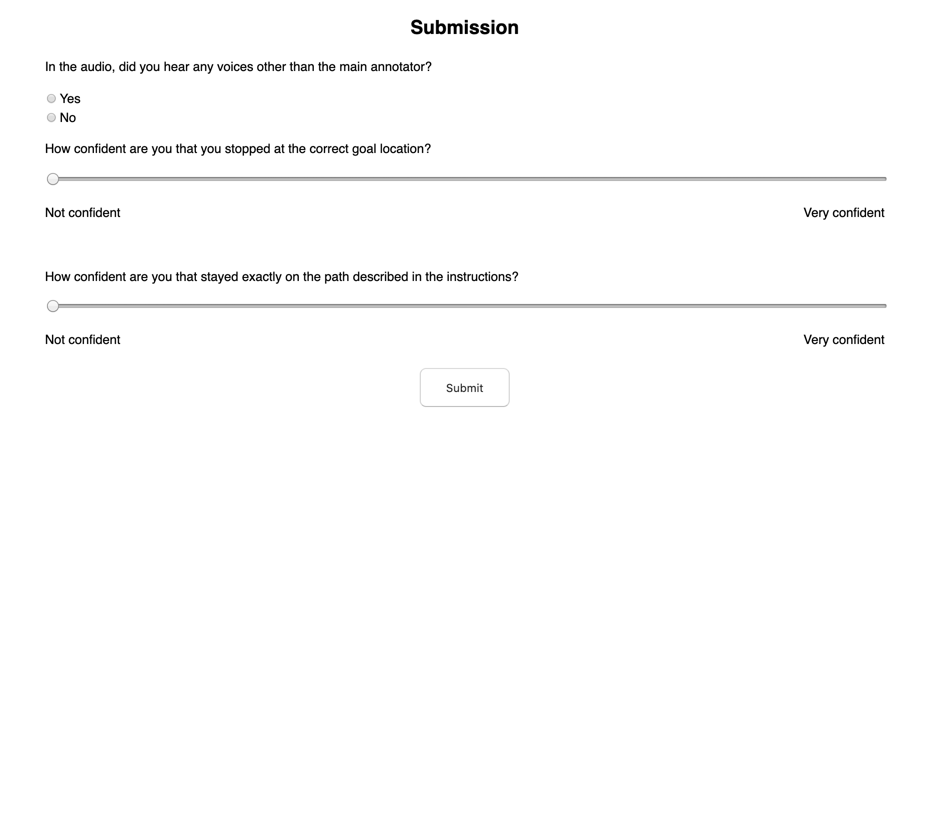}} \\
    \small (e) Guide Transcription (358.3 seconds) & \small (f) Follower Survey (15.2 seconds) \\[6pt]
\end{tabularx}
\end{center}
\caption{Screenshots of the Guide (a, c, e) and Follower (b, d, f) views in our annotation tool, and the average duration for each phase during collection of the first 33K instructions.}
\label{fig:plugins}
\end{figure*}

\begin{figure*}
\centering 
\scriptsize
\begin{tabularx}{\linewidth}{X X X}
\multicolumn{1}{c}{\normalsize \textbf{Guide Alignment}} & \multicolumn{1}{c}{\small ordered left-to-right $\rightarrow$} \\[1mm]
\includegraphics[width=0.32\textwidth]{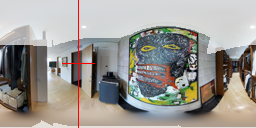} & 
\includegraphics[width=0.32\textwidth]{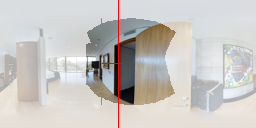} & 
\includegraphics[width=0.32\textwidth]{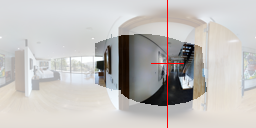} \\
You're starting in a closet, facing an abstract painting on your right. Just slightly to your left will be an open, wooden door next to an amp. Walk through that wooden door. &
&
This will take you to a hallway with stairs going up on the right hand side. Just go straight down the hallway... \\[1mm]
\includegraphics[width=0.32\textwidth]{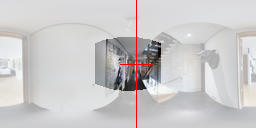} & 
\includegraphics[width=0.32\textwidth]{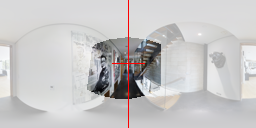} & 
\includegraphics[width=0.32\textwidth]{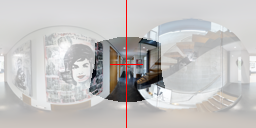} \\
...about five steps... &
...steps. You're going to pass the... &
...the stairs. \\[1mm]
\includegraphics[width=0.32\textwidth]{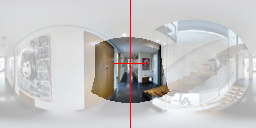} & 
\includegraphics[width=0.32\textwidth]{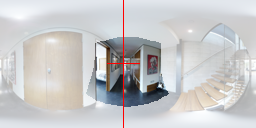} & 
\includegraphics[width=0.32\textwidth]{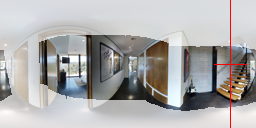} \\
Go one... &
...one step past the stairs. You'll just pass the Albert Einstein painting on your right, and an open doorway on your left. &
There will be a guitar on the floor. At this point, turn around and go up the stairs. \\[1mm]
\includegraphics[width=0.32\textwidth]{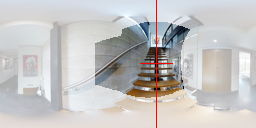} & 
\includegraphics[width=0.32\textwidth]{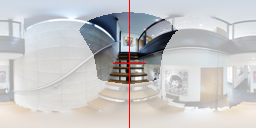} & 
\includegraphics[width=0.32\textwidth]{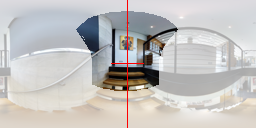} \\
&
&
\\[1mm]
\includegraphics[width=0.32\textwidth]{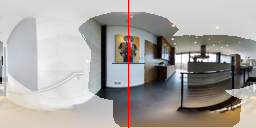} & 
\includegraphics[width=0.32\textwidth]{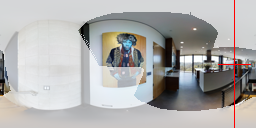} & 
\includegraphics[width=0.32\textwidth]{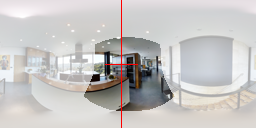} \\
Once you get to the Jimi Hendrix painting, turn... &
...turn to your right and walk between the stair railing and the white kitchen cabinet toward... &
...toward the refrigerator. Take a step in front of the refrigerator.\\[1mm]
\includegraphics[width=0.32\textwidth]{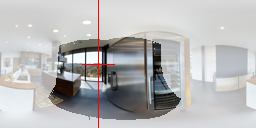} & 
\includegraphics[width=0.32\textwidth]{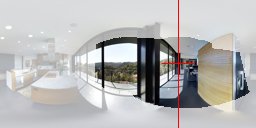} & 
\includegraphics[width=0.32\textwidth]{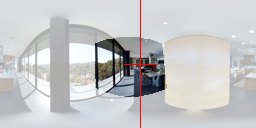} \\
Take another step toward the windows overlooking the trees. &
Then take a right at the end of the refrigerator. You'll take three steps... &
...toward the fireplace.\\[1mm]
\includegraphics[width=0.32\textwidth]{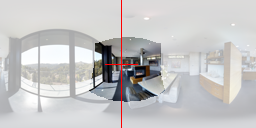} & 
\includegraphics[width=0.32\textwidth]{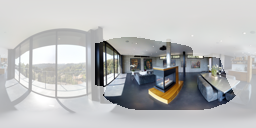} &  \\
Once you get... &
...get to the fireplace, it will be on your right hand side. This is where you stop. &\\
\end{tabularx}
\caption{Spatiotemporal alignment of textual instructions, visual percepts and actions for a long (19-step) en-US \textbf{Guide} path. The next action is indicated in red and unseen pixels in the panoramic images are faded. }
\label{fig:long-guide-trace}
\end{figure*}

\begin{figure*}
\centering 
\scriptsize
\begin{tabularx}{\linewidth}{X X X}
\multicolumn{1}{c}{\normalsize \textbf{Follower Alignment}} & \multicolumn{1}{c}{\small ordered left-to-right $\rightarrow$} & \\[1mm]
\includegraphics[width=0.32\textwidth]{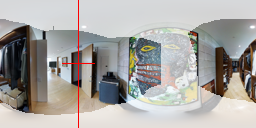} & 
\includegraphics[width=0.32\textwidth]{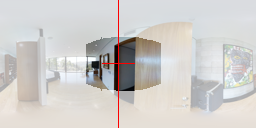} & 
\includegraphics[width=0.32\textwidth]{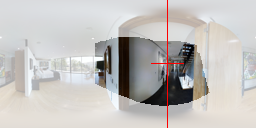} \\
You're starting in a closet, facing an abstract painting on your right. Just slightly to your left will be an open, wooden door next to an amp. Walk through that wooden door.  &
&
This will take you to a hallway with stairs going up on the right hand side. Just... \\[1mm]
\includegraphics[width=0.32\textwidth]{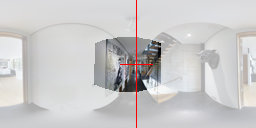} & 
\includegraphics[width=0.32\textwidth]{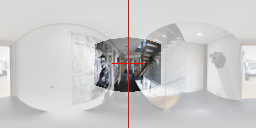} & 
\includegraphics[width=0.32\textwidth]{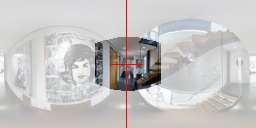} \\
Just go straight down the hallway... &
...about five steps. &
You're going to pass the stairs. \\[1mm]
\includegraphics[width=0.32\textwidth]{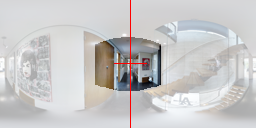} & 
\includegraphics[width=0.32\textwidth]{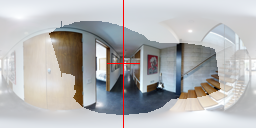} & 
\includegraphics[width=0.32\textwidth]{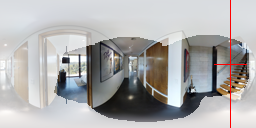} \\
...stairs. &
Go one step past the stairs. You'll just pass the Albert Einstein painting on your right, and an open doorway... &
...doorway on your left. There will be a guitar on the floor. At this point, turn around and go up the stairs.  \\[1mm]
\includegraphics[width=0.32\textwidth]{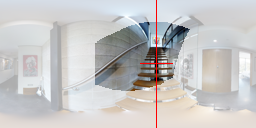} & 
\includegraphics[width=0.32\textwidth]{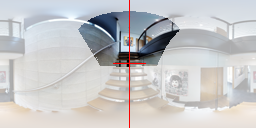} & 
\includegraphics[width=0.32\textwidth]{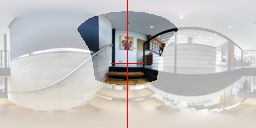} \\
&
Once you get to the... &
Jimi Hendrix painting, turn to your right and...  \\[1mm]
\includegraphics[width=0.32\textwidth]{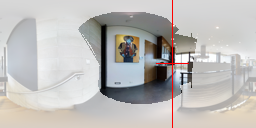} & 
\includegraphics[width=0.32\textwidth]{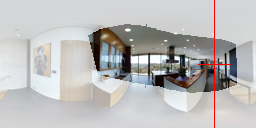} & 
\includegraphics[width=0.32\textwidth]{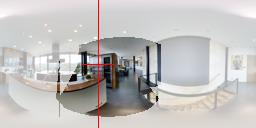} \\
...and walk between the stair railing and the white kitchen cabinet toward the refrigerator. &
Take a step in front of the refrigerator. &
Take another step toward the windows... \\[1mm]
\includegraphics[width=0.32\textwidth]{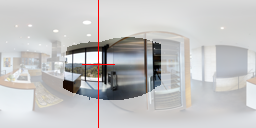} & 
\includegraphics[width=0.32\textwidth]{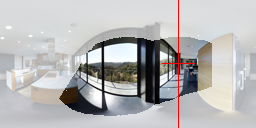} & 
\includegraphics[width=0.32\textwidth]{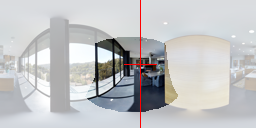} \\
...windows overlooking the trees. &
Then take a right at the end of the refrigerator. You'll take three steps toward the fireplace. &
...fireplace. Once you get to the fireplace, it will be on your right hand side. \\[1mm]
\includegraphics[width=0.32\textwidth]{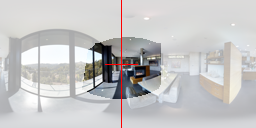} & 
\includegraphics[width=0.32\textwidth]{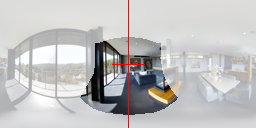} & 
\includegraphics[width=0.32\textwidth]{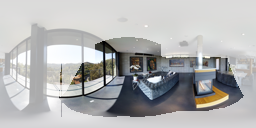}\\
...side. This is where you stop. &
&
\\
\end{tabularx}
\caption{Spatiotemporal alignment of textual instructions, visual percepts and actions for a long en-US \textbf{Follower} path. The next action is indicated in red and unseen pixels in the panoramic images are faded. }
\label{fig:long-follower-trace}
\end{figure*}

\begin{figure*}[t]
\begin{center}
\includegraphics[width=1\linewidth]{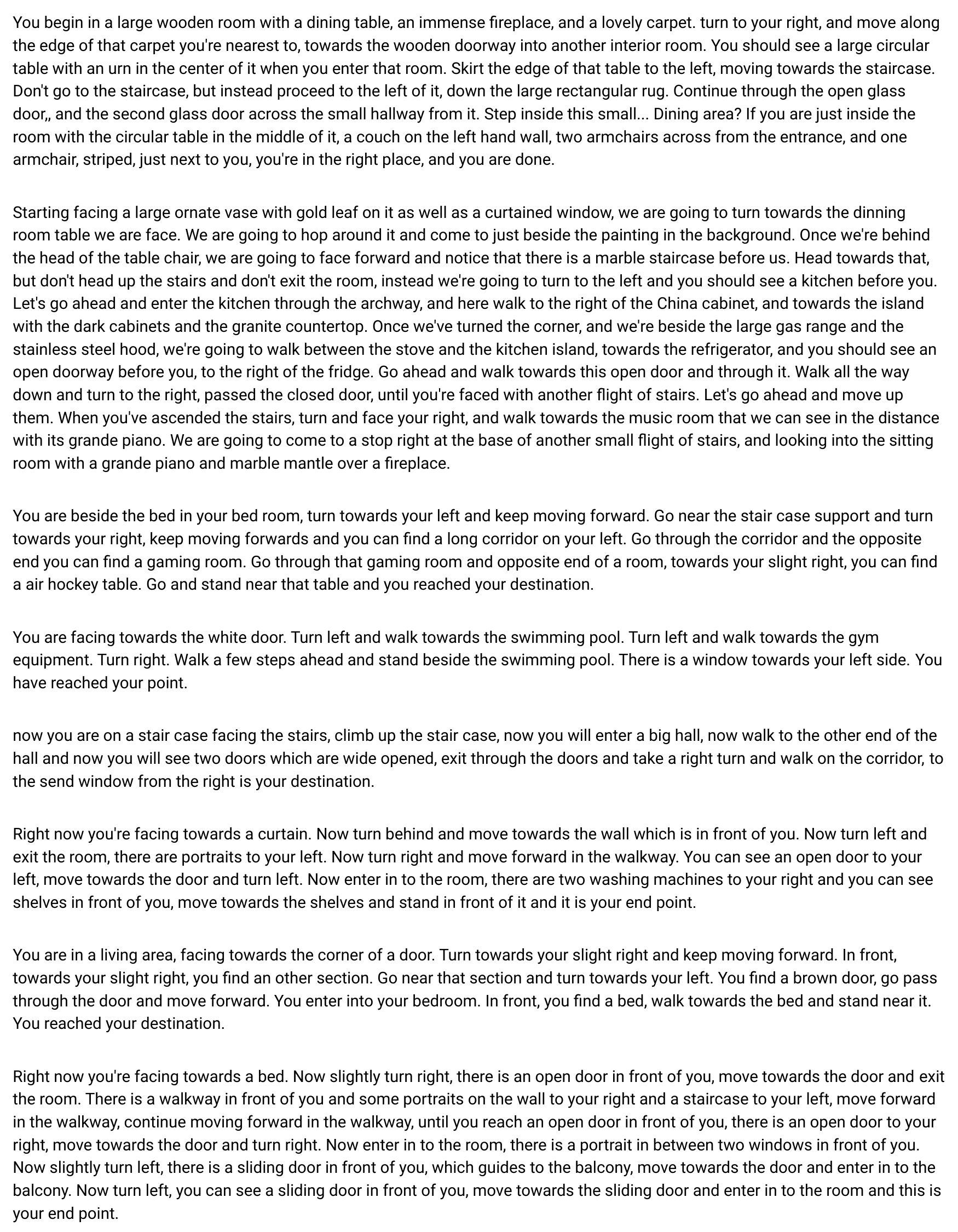}
\end{center}
\caption{Randomly selected English navigation instructions from RxR train. The first two examples are US English and the others are Indian English.}
\label{fig:english}
\end{figure*}

\begin{figure*}[t]
\begin{center}
\includegraphics[width=1\linewidth]{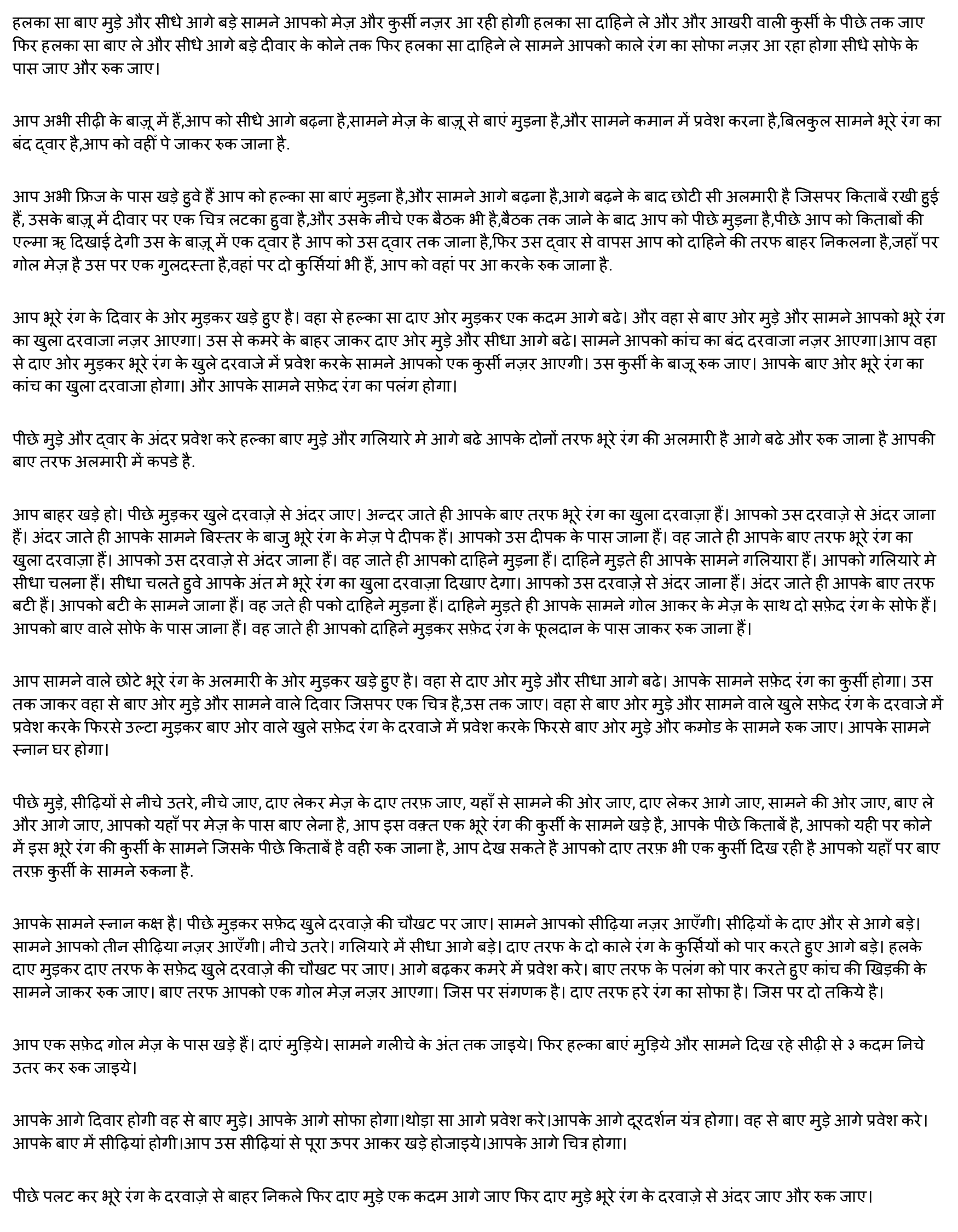}
\end{center}
\caption{Randomly selected Hindi navigation instructions from RxR train.}
\label{fig:hindi}
\end{figure*}

\begin{figure*}[t]
\begin{center}
\includegraphics[width=1\linewidth]{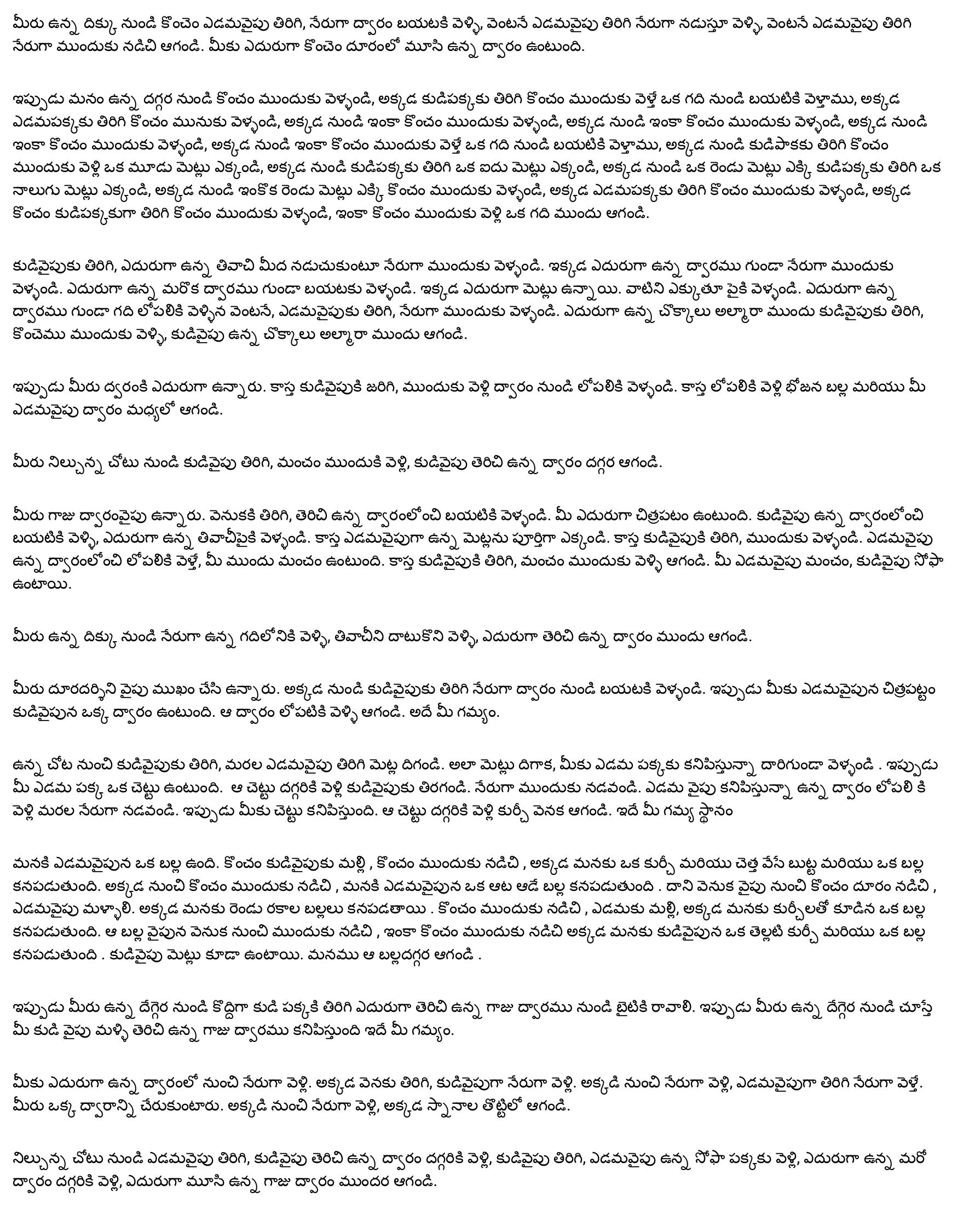}
\end{center}
\caption{Randomly selected Telugu navigation instructions from RxR train.}
\label{fig:telugu}
\end{figure*}

\clearpage

\section{DATASHEET: ROOM-ACROSS-ROOM (RxR)}
\label{sec:datasheet}

%by Gebru \textit{et al.}
\textbf{
This document is based on \textit{Datasheets for Datasets} \cite{gebruDatasheetsDatasets2020}. Please see the most updated version
\underline{\textcolor{blue}{\href{http://arxiv.org/abs/1803.09010}{here}}}.
}

%%%%%%%%%%%%%%%%%%%%%%%%%%%%%%%%%%%%%%%%%%%%%%%%%%%%%%%%%%%%%%%%%%%%%%%%%%%%%%%%
%\begin{mdframed}
\begin{mdframed}[linecolor=\sectioncolor]
\section*{\textcolor{\sectioncolor}{
    MOTIVATION
}}
\end{mdframed}

    \textcolor{\sectioncolor}{\textbf{
    For what purpose was the dataset created?
    }
    %Was there a specific task in mind? Was there
    %a specific gap that needed to be filled? Please provide a description.
    } \\
    %%%
    RxR was created to advance progress on vision-and-language navigation (VLN) in multiple languages (English, Hindi, Telugu). It addresses gaps in existing datasets by including more paths that counter known biases and an order of magnitude more navigation instructions for three languages plus annotators' 3D virtual pose sequences. \\
    %%% 
    
    \textcolor{\sectioncolor}{\textbf{
    Who created this dataset (e.g., which team, research group) and on behalf
    of which entity (e.g., company, institution, organization)?
    }
    } \\
    %%%
    This dataset was created by Alexander Ku, Peter Anderson, Roma Patel, Eugene Ie, Jason Baldridge and the Google Data Compute team on behalf of Google Research. \\
    %%% 
    
    \textcolor{\sectioncolor}{\textbf{
    What support was needed to make this dataset?
    }
    %(e.g.who funded the creation of the dataset? If there is an associated
    %grant, provide the name of the grantor and the grant name and number, or if
    %it was supported by a company or government agency, give those details.)
    } \\
    %%%
    Funding was provided by Google Research. \\
    %%% 
    
    %\textcolor{\sectioncolor}{\textbf{
    %Any other comments?
    %}} \\
    %%%
    %None. \\
    %%%

%%%%%%%%%%%%%%%%%%%%%%%%%%%%%%%%%%%%%%%%%%%%%%%%%%%%%%%%%%%%%%%%%%%%%%%%%%%%%%%%
\begin{mdframed}[linecolor=\sectioncolor]
\section*{\textcolor{\sectioncolor}{
    COMPOSITION
}}
\end{mdframed}
    \textcolor{\sectioncolor}{\textbf{
    What do the instances that comprise the dataset represent (e.g., documents,
    photos, people, countries)?
    }
    %Are there multiple types of instances (e.g., movies, users, and ratings;
    %people and interactions between them; nodes and edges)? Please provide a
    %description.
    } \\
    %%%
    The instances in RxR are natural language navigation instructions paired with trajectories in reconstructed 3D buildings. Each navigation instruction has been recorded as speech and transcribed by the speaker. The dataset includes the text transcriptions, but not the audio files, although they may be released in future. The trajectories are provided as paths, consisting of sequences of viewpoint ids corresponding to navigation graphs from \citet{mattersim}, and pose traces, consisting of sequences of virtual camera poses situated in the underlying building reconstructions which are from the Matterport3D dataset \cite{Matterport3D}. Pose traces and text transcriptions are timestamped and aligned. Pose traces are provided for both the instruction annotator (the Guide), and a second annotator charged with following the Guide's instructions (the Follower). \\
    %%% 
    
    \textcolor{\sectioncolor}{\textbf{
    How many instances are there in total (of each type, if appropriate)?
    }
    } \\
    %%%
    RxR contains 126K Guide instructions covering 16.5K sampled paths and 126K human Follower demonstration paths. Annotations are split equally across the three languages in the dataset.
    Refer to Table \ref{tab:datasets} for a comparison of the number of instances to previous datasets and Table \ref{tab:summary} for summary statistics. 
     \\
    %%% 
    
    \textcolor{\sectioncolor}{\textbf{
    Does the dataset contain all possible instances or is it a sample (not
    necessarily random) of instances from a larger set?
    }
    %If the dataset is a sample, then what is the larger set? Is the sample
    %representative of the larger set (e.g., geographic coverage)? If so, please
    %describe how this representativeness was validated/verified. If it is not
    %representative of the larger set, please describe why not (e.g., to cover a
    %more diverse range of instances, because instances were withheld or
    %unavailable).
    } \\
    %%%
    Refer to Section \ref{sec:sampling} for a detailed description of the sampling procedure used to select the paths for annotation.\\
    %%% 
    
    \textcolor{\sectioncolor}{\textbf{
    What data does each instance consist of?
    }
    %“Raw” data (e.g., unprocessed text or images) or features? In either case,
    %please provide a description.
    } \\
    %%%
    Each instance consists of a trajectory through a building from the Matterport3D dataset \cite{Matterport3D} paired with a natural language navigation instruction. A trajectory can be visualized as a sequence of 360-degree panoramic images, or as path traversing a 3D reconstruction of the building represented as a textured mesh. Refer to Table \ref{tab:phenomena} for an analysis of linguistic phenomena in the instructions and Figures \ref{fig:english}, \ref{fig:hindi} and \ref{fig:telugu} for instruction examples in English, Hindi and Telugu respectively. \\
    %%% 
    
    \textcolor{\sectioncolor}{\textbf{
    Is there a label or target associated with each instance?
    }
    %If so, please provide a description.
    } \\
    %%%
    When training wayfinding agents to navigate from natural language instructions, the trajectory is the target. Instructions and paths are annotated with unique identifiers. 
    \\
    %%% 
    
    \textcolor{\sectioncolor}{\textbf{
    Is any information missing from individual instances?
    }
    %If so, please provide a description, explaining why this information is
    %missing (e.g., because it was unavailable). This does not include
    %intentionally removed information, but might include, e.g., redacted text.
    } \\
    %%%
    We do not provide the Guide audio recordings, for reasons outlined in Appendix A. \\
    %%% 
    
    \textcolor{\sectioncolor}{\textbf{
    Are relationships between individual instances made explicit (e.g., users’
    movie ratings, social network links)?
    }
    %If so, please describe how these relationships are made explicit.
    } \\
    %%%
    Trajectories may belong to the same building or different buildings; each instance is annotated with a scan (building) identifier. \\
    %%% 
    
    \textcolor{\sectioncolor}{\textbf{
    Are there recommended data splits (e.g., training, development/validation,
    testing)?
    }
    %If so, please provide a description of these splits, explaining the
    %rationale behind them.
    } \\
    %%%
    Yes. We follow the same building splits as Matterport3D and R2R. Refer to Section \ref{par:path-stats} for details regarding the RxR train/validation/test instance splits. \\
    %%% 
    
    \textcolor{\sectioncolor}{\textbf{
    Are there any errors, sources of noise, or redundancies in the dataset?
    }
    %If so, please provide a description.
    } \\
    %%%
    The process we followed to validate instruction quality using Follower annotations is described in Section \ref{sec:follower}. \\
    %%% 
    
    \textcolor{\sectioncolor}{\textbf{
    Is the dataset self-contained, or does it link to or otherwise rely on
    external resources (e.g., websites, tweets, other datasets)?
    }
    %If it links to or relies on external resources, a) are there guarantees
    %that they will exist, and remain constant, over time; b) are there official
    %archival versions of the complete dataset (i.e., including the external
    %resources as they existed at the time the dataset was created); c) are
    %there any restrictions (e.g., licenses, fees) associated with any of the
    %external resources that might apply to a future user? Please provide
    %descriptions of all external resources and any restrictions associated with
    %them, as well as links or other access points, as appropriate.
    } \\
    %%%
    This dataset is based on building reconstructions from the Matterport3D dataset \cite{Matterport3D} and viewpoint navigation graphs from the R2R dataset \cite{mattersim}. Apart from these dependencies, RxR is self-contained, i.e., it does not rely on web resources. \\
    %%% 
    
    \textcolor{\sectioncolor}{\textbf{
    Does the dataset contain data that might be considered confidential (e.g.,
    data that is protected by legal privilege or by doctor-patient
    confidentiality, data that includes the content of individuals’ non-public
    communications)?
    }
    %If so, please provide a description.
    } \\
    %%%
    No. \\
    %%% 
    
    \textcolor{\sectioncolor}{\textbf{
    Does the dataset contain data that, if viewed directly, might be offensive,
    insulting, threatening, or might otherwise cause anxiety?
    }
    %If so, please describe why.
    } \\
    %%%
    No. \\
    %%% 
    
    %\textcolor{\sectioncolor}{\textbf{
    %Does the dataset relate to people?
    %}
    %If not, you may skip the remaining questions in this section.
    %} \\
    %%%
    %No. \\
    %%% 
    
    \textcolor{\sectioncolor}{\textbf{
    Does the dataset identify any subpopulations (e.g., by age, gender)?
    }
    %If so, please describe how these subpopulations are identified and
    %provide a description of their respective distributions within the dataset.
    } \\
    %%%
    No. \\
    %%% 
    
    \textcolor{\sectioncolor}{\textbf{
    Is it possible to identify individuals (i.e., one or more natural persons),
    either directly or indirectly (i.e., in combination with other data) from
    the dataset?
    }
    %If so, please describe how.
    } \\
    %%%
    No. \\
    %%% 
    
    \textcolor{\sectioncolor}{\textbf{
    Does the dataset contain data that might be considered sensitive in any way
    (e.g., data that reveals racial or ethnic origins, sexual orientations,
    religious beliefs, political opinions or union memberships, or locations;
    financial or health data; biometric or genetic data; forms of government
    identification, such as social security numbers; criminal history)?
    }
    %If so, please provide a description.
    } \\
    %%%
    Each natural language instruction in the sample is either in English, Hindi or Telugu, thus potentially revealing linguistic origin. However, no other annotator data is included in the dataset. \\
    %%% 
    
    %\textcolor{\sectioncolor}{\textbf{
    %Any other comments?
    %}} \\
    %%%
    %None. \\
    %%%

%%%%%%%%%%%%%%%%%%%%%%%%%%%%%%%%%%%%%%%%%%%%%%%%%%%%%%%%%%%%%%%%%%%%%%%%%%%%%%%%
\begin{mdframed}[linecolor=\sectioncolor]
\section*{\textcolor{\sectioncolor}{
    COLLECTION
}}
\end{mdframed}

    \textcolor{\sectioncolor}{\textbf{
    How was the data associated with each instance acquired?
    }
    %Was the data directly observable (e.g., raw text, movie ratings),
    %reported by subjects (e.g., survey responses), or indirectly
    %inferred/derived from other data (e.g., part-of-speech tags, model-based
    %guesses for age or language)? If data was reported by subjects or
    %indirectly inferred/derived from other data, was the data
    %validated/verified? If so, please describe how.
    } \\
    %%%
    Refer to Section \ref{sec:collection} for details of the annotation procedure, as well as measures undertaken to validate the data. \\
    %%% 
    
    \textcolor{\sectioncolor}{\textbf{
    Over what timeframe was the data collected?
    }
    %Does this timeframe match the creation timeframe of the data associated
    %with the instances (e.g., recent crawl of old news articles)? If not,
    %please describe the timeframe in which the data associated with the
    %instances was created. Finally, list when the dataset was first published.
    } \\
    %%%
    The dataset was collected between March 2020 and September 2020.  \\
    %%% 
    
    \textcolor{\sectioncolor}{\textbf{
    What mechanisms or procedures were used to collect the data (e.g., hardware
    apparatus or sensor, manual human curation, software program, software
    API)?
    }
    %How were these mechanisms or procedures validated?
    } \\
    %%%
    We developed a web-based annotation tool to collect the data. It is described further in Section \ref{sec:collection} and screenshots are included in Figure \ref{fig:plugins}. \\
    %%% 
    
    %\textcolor{\sectioncolor}{\textbf{
    %What was the resource cost of collecting the data?
    %}
    %(e.g. what were the required computational resources, and the associated
    %financial costs, and energy consumption - estimate the carbon footprint.
    %See Strubell \textit{et al.}\cite{strubellEnergyPolicyConsiderations2019} for approaches in this area.)
    %} \\
    %%%
    %\textbf{TODO: cost}. \\
    %%% 
    
    \textcolor{\sectioncolor}{\textbf{
    If the dataset is a sample from a larger set, what was the sampling
    strategy (e.g., deterministic, probabilistic with specific sampling
    probabilities)?
    }
    } \\
    %%%
    Please see Section \ref{sec:sampling} and Figure \ref{fig:sampling} for details of the strategy for selecting paths for annotation. \\
    %%% 
    
    \textcolor{\sectioncolor}{\textbf{
    Who was involved in the data collection process (e.g., students,
    crowdworkers, contractors) and how were they compensated (e.g., how much
    were crowdworkers paid)?
    }
    } \\
    %%%
    Refer to Appendix A. \\
    %%% 
    
    %\textcolor{\sectioncolor}{\textbf{
    %Were any ethical review processes conducted (e.g., by an institutional
    %review board)?
    %}
    %If so, please provide a description of these review processes, including
    %the outcomes, as well as a link or other access point to any supporting
    %documentation.
    %} \\
    %%%
    %\textbf{TODO..} \\
    %%% 
    
    \textcolor{\sectioncolor}{\textbf{
    Does the dataset relate to people?
    }
    %If not, you may skip the remainder of the questions in this section.
    } \\
    %%%
    Yes. \\
    %%% 
    
    \textcolor{\sectioncolor}{\textbf{
    Did you collect the data from the individuals in question directly, or
    obtain it via third parties or other sources (e.g., websites)?
    }
    } \\
    %%%
    Directly from the individuals. \\
    %%% 
    
    \textcolor{\sectioncolor}{\textbf{
    Were the individuals in question notified about the data collection?
    }
    %If so, please describe (or show with screenshots or other information) how
    %notice was provided, and provide a link or other access point to, or
    %otherwise reproduce, the exact language of the notification itself.
    } \\
    %%%
    Yes.  \\
    %%% 
    
    \textcolor{\sectioncolor}{\textbf{
    Did the individuals in question consent to the collection and use of their
    data?
    }
    %If so, please describe (or show with screenshots or other information) how
    %consent was requested and provided, and provide a link or other access
    %point to, or otherwise reproduce, the exact language to which the
    %individuals consented.
    } \\
    %%%
    Yes.  \\
    %%% 
    
    %\textcolor{\sectioncolor}{\textbf{
    %If consent was obtained, were the consenting individuals provided with a
    %mechanism to revoke their consent in the future or for certain uses?
    %}
    % If so, please provide a description, as well as a link or other access
    % point to the mechanism (if appropriate)
    %} \\
    %%%
    %No. \\
    %%% 
    
    %\textcolor{\sectioncolor}{\textbf{
    %Has an analysis of the potential impact of the dataset and its use on data
    %subjects (e.g., a data protection impact analysis)been conducted?
    %}
    %If so, please provide a description of this analysis, including the
    %outcomes, as well as a link or other access point to any supporting
    %documentation.
    %} \\
    %%%
    %No. \\
    %%% 
    
    %\textcolor{\sectioncolor}{\textbf{
    %Any other comments?
    %}} \\
    %%%
    %None. \\
    %%%

%%%%%%%%%%%%%%%%%%%%%%%%%%%%%%%%%%%%%%%%%%%%%%%%%%%%%%%%%%%%%%%%%%%%%%%%%%%%%%%%
\begin{mdframed}[linecolor=\sectioncolor]
\section*{\textcolor{\sectioncolor}{
    PREPROCESSING / CLEANING / LABELING
}}
\end{mdframed}

    \textcolor{\sectioncolor}{\textbf{
    Was any preprocessing/cleaning/labeling of the data
    done(e.g.,discretization or bucketing, tokenization, part-of-speech
    tagging, SIFT feature extraction, removal of instances, processing of
    missing values)?
    }
   % If so, please provide a description. If not, you may skip the remainder of
    %the questions in this section.
    } \\
    %%%
    Please see Section \ref{sec:follower} describing the process used to remove and re-annotate instances in which the Follower was not able to correctly follow the path described in the Guide's instruction. \\
    %%%

    \textcolor{\sectioncolor}{\textbf{
    Was the “raw” data saved in addition to the preprocessed/cleaned/labeled
    data (e.g., to support unanticipated future uses)?
    }
    %If so, please provide a link or other access point to the “raw” data.
    } \\
    %%%
    Yes. \\
    %%%

    \textcolor{\sectioncolor}{\textbf{
    Is the software used to preprocess/clean/label the instances available?
    }
    %If so, please provide a link or other access point.
    } \\
    %%%
    We plan to publicly release our web-based annotation tool. \\
    %%%

    %\textcolor{\sectioncolor}{\textbf{
    %Any other comments?
    %}} \\
    %%%
    %None. \\
    %%%

%%%%%%%%%%%%%%%%%%%%%%%%%%%%%%%%%%%%%%%%%%%%%%%%%%%%%%%%%%%%%%%%%%%%%%%%%%%%%%%%
\begin{mdframed}[linecolor=\sectioncolor]
\section*{\textcolor{\sectioncolor}{
    USES
}}
\end{mdframed}

    \textcolor{\sectioncolor}{\textbf{
    Has the dataset been used for any tasks already?
    }
    %If so, please provide a description.
    } \\
    %%%
    We have used RxR to train vision-and-language navigation (VLN) agents as described in the paper.  \\
    %%%

    \textcolor{\sectioncolor}{\textbf{
    Is there a repository that links to any or all papers or systems that use the dataset?
    }
    %If so, please provide a link or other access point.
    } \\
    %%%
    No, although we plan to release a test server and leaderboard to support the research community using the dataset. \\
    %%%

    \textcolor{\sectioncolor}{\textbf{
    What (other) tasks could the dataset be used for?
    }
    } \\
    %%%
    Training models to generate natural language navigation instructions, visual referring expression grounding and comprehension, grounded dialog tasks, pre-training for various other vision-and-language tasks, multilingual learning and so on. \\
    %%%

    \textcolor{\sectioncolor}{\textbf{
    Is there anything about the composition of the dataset or the way it was
    collected and preprocessed/cleaned/labeled that might impact future uses?
    }
    %For example, is there anything that a future user might need to know to
    %avoid uses that could result in unfair treatment of individuals or groups
    %(e.g., stereotyping, quality of service issues) or other undesirable harms
    %(e.g., financial harms, legal risks) If so, please provide a description.
    %Is there anything a future user could do to mitigate these undesirable
    %harms?
    } \\
    %%%
    No. \\
    %%%

    %\textcolor{\sectioncolor}{\textbf{
    %Are there tasks for which the dataset should not be used?
    %}
    %If so, please provide a description.
    %} \\
    %%%
    %No. \\
    %%%

    %\textcolor{\sectioncolor}{\textbf{
    %Any other comments?
    %}} \\
    %%%
    %None. \\
    %%%

%%%%%%%%%%%%%%%%%%%%%%%%%%%%%%%%%%%%%%%%%%%%%%%%%%%%%%%%%%%%%%%%%%%%%%%%%%%%%%%%
\begin{mdframed}[linecolor=\sectioncolor]
\section*{\textcolor{\sectioncolor}{
    DISTRIBUTION
}}
\end{mdframed}

    \textcolor{\sectioncolor}{\textbf{
    Will the dataset be distributed to third parties outside of the entity
    (e.g., company, institution, organization) on behalf of which the dataset
    was created?
    }
    %If so, please provide a description.
    } \\
    %%%
    Yes, this dataset is open to use by the research community. \\
    %%%

    \textcolor{\sectioncolor}{\textbf{
    How will the dataset will be distributed (e.g., tarball on website, API,
    GitHub)?
    }
    %Does the dataset have a digital object identifier (DOI)?
    } \\
    %%%
    Via GitHub and Google Cloud Storage. \\
    %%%

    \textcolor{\sectioncolor}{\textbf{
    When will the dataset be distributed?
    }
    } \\
    %%%
    This dataset has been distributed on publication. \\
    %%%

    \textcolor{\sectioncolor}{\textbf{
    Will the dataset be distributed under a copyright or other intellectual
    property (IP) license, and/or under applicable terms of use (ToU)?
    }
    %If so, please describe this license and/or ToU, and provide a link or other
    %access point to, or otherwise reproduce, any relevant licensing terms or
    %ToU, as well as any fees associated with these restrictions.
    } \\
    %%%
    RxR is released under a \href{https://creativecommons.org/licenses/by/4.0/}{CC-BY license}. \\
    %%%

    \textcolor{\sectioncolor}{\textbf{
    Have any third parties imposed IP-based or other restrictions on the data
    associated with the instances?
    }
    %If so, please describe these restrictions, and provide a link or other
    %access point to, or otherwise reproduce, any relevant licensing terms, as
    %well as any fees associated with these restrictions.
    } \\
    %%%
    Yes, the Matterport3D dataset is governed by the \href{http://kaldir.vc.in.tum.de/matterport/MP_TOS.pdf}{Matterport3D Terms of Use}. \\
    %%%

    %\textcolor{\sectioncolor}{\textbf{
    %Do any export controls or other regulatory restrictions apply to the
    %dataset or to individual instances?
    %}
    %If so, please describe these restrictions, and provide a link or other
    %access point to, or otherwise reproduce, any supporting documentation.
    %} \\
    %%%
    %No. \\
    %%%

    %\textcolor{\sectioncolor}{\textbf{
    %Any other comments?
    %}} \\
    %%%
    %None. \\
    %%%

%%%%%%%%%%%%%%%%%%%%%%%%%%%%%%%%%%%%%%%%%%%%%%%%%%%%%%%%%%%%%%%%%%%%%%%%%%%%%%%%
\begin{mdframed}[linecolor=\sectioncolor]
\section*{\textcolor{\sectioncolor}{
    MAINTENANCE
}}
\end{mdframed}

    %\textcolor{\sectioncolor}{\textbf{
    %Who is supporting/hosting/maintaining the dataset?
    %}
    %} \\
    %%%
    %The dataset has been released on publication and will be hosted at the GitHub repository here. \textbf{TODO exact link}. \\
    %%%

    \textcolor{\sectioncolor}{\textbf{
    How can the owner/curator/manager of the dataset be contacted (e.g., email
    address)?
    }
    } \\
    %%%
    Email contact: \texttt{rxrvln@google.com}. \\
    %%%

    %\textcolor{\sectioncolor}{\textbf{
    %Is there an erratum?
    %}
    %If so, please provide a link or other access point.
    %} \\
    %%%
    %No. \\
    %%%

    \textcolor{\sectioncolor}{\textbf{
    Will the dataset be updated (e.g., to correct labeling errors, add new
    instances, delete instances)?
    }
    %If so, please describe how often, by whom, and how updates will be
    %communicated to users (e.g., mailing list, GitHub)?
    } \\
    %%%
    No. \\
    %%%

    %\textcolor{\sectioncolor}{\textbf{
    %If the dataset relates to people, are there applicable limits on the
    %retention of the data associated with the instances (e.g., were individuals
    %in question told that their data would be retained for a fixed period of
    %time and then deleted)?
    %}
    %If so, please describe these limits and explain how they will be enforced.
    %} \\
    %%%
    %No. \\
    %%%

    %\textcolor{\sectioncolor}{\textbf{
    %Will older versions of the dataset continue to be
    %supported/hosted/maintained?
    %}
    %If so, please describe how. If not, please describe how its obsolescence
    %will be communicated to users.
    %} \\
    %%%
    %N/A. \\
    %%%

    %\textcolor{\sectioncolor}{\textbf{
    %If others want to extend/augment/build on/contribute to the dataset, is
    %there a mechanism for them to do so?
    %}
    %If so, please provide a description. Will these contributions be
    %validated/verified? If so, please describe how. If not, why not? Is there a
    %process for communicating/distributing these contributions to other users?
    %If so, please provide a description.
    %} \\
    %%%
    %Researchers are welcome to augment and extend the dataset as they see fit for certain research directions e.g., more languages for instructions, finer-grained visual annotations and so on. \\
    %%%

    %\textcolor{\sectioncolor}{\textbf{
    %Any other comments?
    %}} \\
    %%%
    %None. \\
    %%%

\end{document}